\documentclass{article} %
\usepackage[T1]{fontenc} %
\PassOptionsToPackage{table}{xcolor} %
\usepackage{preprint,times}

\usepackage{amsmath}
\usepackage{amssymb}
\usepackage{booktabs}
\usepackage{float}
\usepackage{placeins}
\usepackage{graphicx}
\graphicspath{{figures/}}
\usepackage{multirow}
\usepackage{array}
\usepackage{url}
\usepackage{hyperref}
\usepackage{tikz}
\usetikzlibrary{positioning,arrows.meta,fit}
\usepackage{pgfplots}
\pgfplotsset{compat=1.17}

\definecolor{ourrow}{RGB}{223,235,255} %
\definecolor{ourvariantrow}{RGB}{241,246,255} %
\definecolor{losered}{RGB}{252,222,222} %
\definecolor{basegray}{RGB}{240,240,240} %

\newcommand{\ours}{Latent-MOPD}
\newcommand{\contribnum}[1]{\tikz[baseline=(n.base)]{\node[circle,fill=black,text=white,inner sep=0pt,minimum size=8pt,font=\fontsize{6}{6}\selectfont\bfseries] (n) {#1};}}
\newcommand{\lopd}{\mathcal{L}_{\mathrm{OPD}}}
\newcommand{\lrep}{\mathcal{L}_{\mathrm{rep}}}
\newcommand{\tablefont}{\normalfont\fontsize{8.5}{9.5}\selectfont}

\newcommand{\sg}{\mathrm{sg}}
\newcommand{\cmark}{\ding{51}}
\newcommand{\xmark}{\ding{55}}
\usepackage{pifont}

\title{\ours{}: Latent Multi-Teacher \\ On-Policy Distillation}

\author{%
Zhengyu Fang$^{1}$\thanks{Equal contribution.}\quad
Seoyeon Hong$^{2}$\footnotemark[1]\quad
Jie Yang$^{3}$\footnotemark[1]\quad
Muyang Li$^{4}$\quad
Koyoshi Shindo$^{2}$\\
\bfseries Brandon Joseph Lwowski$^{2}$\quad
Jing Li$^{1}$\thanks{Corresponding author.}\\[4pt]
{\normalfont $^{1}$Case Western Reserve University\quad
$^{2}$Zillow Group, Inc.\quad
$^{3}$University of Illinois at Chicago}\\
{\normalfont $^{4}$University of Florida}\\[4pt]
{\normalfont\fontsize{8}{9.5}\selectfont\texttt{zxf177@case.edu}, \texttt{seoyeonh@zillowgroup.com}, \texttt{jyang265@uic.edu}, \texttt{muyang.li@ufl.edu},}\\
{\normalfont\fontsize{8}{9.5}\selectfont\texttt{koyoshis@zillowgroup.com}, \texttt{brandonlw@zillowgroup.com}, \texttt{jxl175@case.edu}}%
}

\ppfinalcopy

\begin{document}

\maketitle
\lhead{Preprint}

\begin{abstract}
On-policy distillation (OPD) trains a student on the responses it generates. Existing
LLM multi-teacher OPD transfers \emph{what} specialists predict through their output distributions.
We introduce \ours{}, to our knowledge the first representation-level multi-teacher OPD
method for LLMs. It integrates existing specialists through both their predictions and the
hidden states used to compute them, without additional teacher training. To coordinate
representation supervision from multiple specialists, we select late-layer targets according
to the teacher--student relationship, bridge unequal hidden widths with a shared projection,
and group updates by domain. Each teacher's supervision gradually shifts from hidden states
to token predictions, with both channels using the same routed specialist.
In our main same-family setting, \ours{} outperforms the token-only, representation-only and
uniform-averaging baselines on all nine benchmarks across math, code and logic. With the same
parameter count as each teacher, the student also surpasses the per-benchmark best teacher on
a majority of these benchmarks.
With larger, separately developed cross-family teachers, \ours{} outperforms both single-channel baselines on all benchmarks. A same-family all-layer representation-only control remains stable with domain-pure updates but collapses when teacher domains are interleaved within an update. Our results show that a single student can integrate capabilities from several specialists through both their output distributions and internal representations.
\end{abstract}

\section{Introduction}
\label{sec:intro}

\begin{figure}[!t]
\centering
\includegraphics[width=\textwidth]{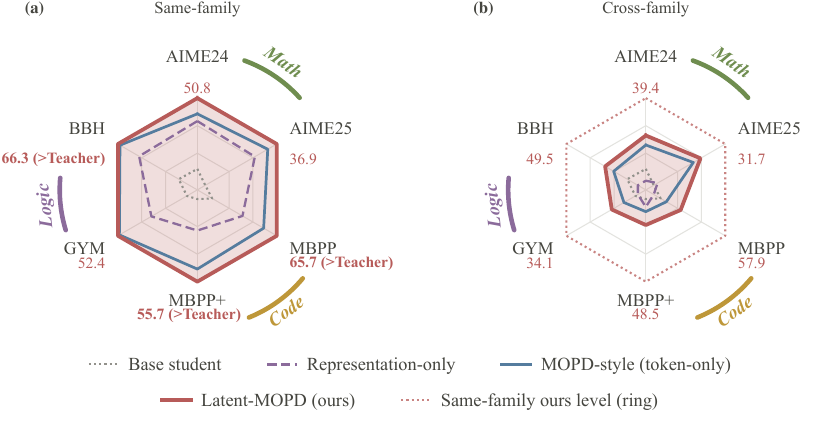}
\caption{\textbf{Same-family gains beyond individual teachers, with a cross-family extension.}
\textbf{(a)} Same-family; \textbf{(b)} cross-family (last 1, Linear). Both panels use shared per-axis scales,
with each outer vertex fixed at the same-family \ours{} score. Solid polygons show \ours{};
the dotted ring in \textbf{(b)} repeats the same-family reference. Representation-only denotes
OPRD-style (all layers) in \textbf{(a)} and Rep-only (last 1) in \textbf{(b)}.
Labels give \ours{} scores; \textbf{bold} with ($>$Teacher) marks those above the strongest teacher.
Table~\ref{tab:s1} reports all nine same-family benchmarks.}
\label{fig:teaser}
\vspace{-8pt}
\end{figure}

Reinforcement learning~(RL) has become an important tool for improving large language models~(LLMs)~\citep{ouyang2022training,shao2024deepseekmath}, and specialized RL pipelines now produce models with complementary strengths in domains such as mathematical reasoning, coding, and logic.
While each of these models excels in its own domain, a broader goal is a single model that combines their strengths by reusing the existing specialists (Figure~\ref{fig:teaser}).
Multi-teacher on-policy distillation~(MOPD)~\citep{mopd2026} extends on-policy distillation~(OPD)~\citep{agarwal2024onpolicy} to several specialists, routing each student-generated response to its domain's teacher.
The selected teacher scores the student's tokens, and its next-token distribution provides dense token-level guidance along the student's own rollout.
In this way, the student learns what each specialist predicts, with supervision defined entirely at the output level.

These predictions, however, are the end result of internal computations.
Interpretability studies show that hidden representations carry intermediate reasoning steps and causally influence later computation, even when these steps are not verbalized~\citep{lindsey2025biology,gurnee2026workspace}.
This suggests that internal representations carry information about \emph{how} a teacher computes its predictions, not only \emph{what} it predicts.
Representation-level OPD methods such as LastOPD and OPRD show that, with suitable layer pairing and training schedules, hidden-state supervision can improve transfer over token-only OPD~\citep{yang2026lastopd,yang2026oprd}.
However, these approaches study learning from a single teacher.
Learning from multiple specialists poses a different challenge: their distinct representation targets must jointly train the same student backbone.
Routing determines whose supervision a prompt receives, but leaves a central question: \textbf{\emph{how should multi-teacher OPD select representation targets and organize their supervision so that one student integrates the capabilities of several specialists?}}

\textbf{Choosing representation targets requires more than matching layer indices.}
Before training, centered kernel alignment~(CKA)~\citep{kornblith2019similarity} between each same-family specialist and the base student remains high through most layers, with larger differences near the output~(Figure~\ref{fig:cka}a).
Across model families, this correspondence is uneven, with a sharp drop in similarity at intermediate depths~(Figure~\ref{fig:cka}b).
These patterns suggest different considerations for alignment.
Within a shared lineage, the common representational structure provides a basis for learning from the specialists' late-layer differences; across lineages, weak correspondence at intermediate depths instead motivates targets with a shared computational role, such as the states directly used for next-token prediction.
The choice of representation targets, therefore, depends on both the relationship between the models and the role of the states being aligned.

\textbf{Teacher routing does not determine how representation supervision should be organized over training.}
In a same-family all-layer representation-only control, mixing samples supervised by different teachers within an optimizer update collapses training within twenty updates, whereas domain-pure batches remain stable~(Figure~\ref{fig:cka}c; Section~\ref{sec:detonate}).
Each sample receives supervision from the same specialist in both settings; the difference is whether several teachers contribute to a single parameter update.
Sustained representation-only supervision is also problematic: in the cross-family setting, it falls below the student's initial performance in aggregate, so how latent supervision is scheduled over training matters as much as how it is batched.

\begin{figure}[!t]
\centering
\includegraphics[width=\textwidth]{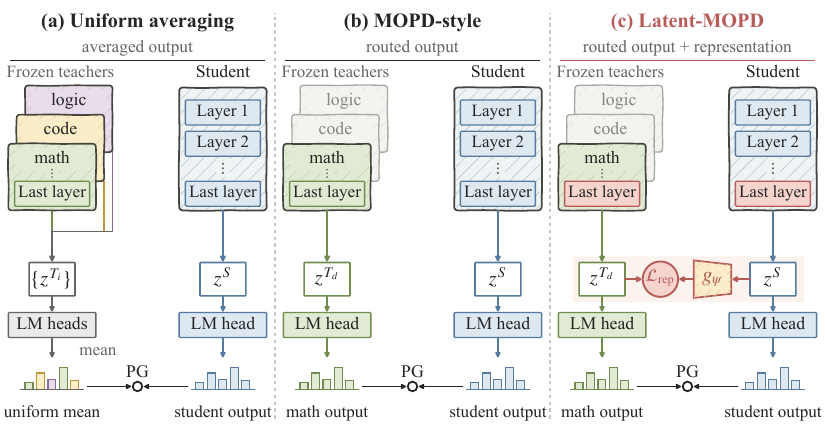}
\caption{\textbf{Three supervision designs on shared student-generated prefixes.}
\textbf{(a)} Uniform averaging gives each teacher's token probabilities equal weight. \textbf{(b)} MOPD-style routes token
supervision to one frozen teacher. \textbf{(c)} \ours{} adds a representation target from
the same teacher, matching $g_\psi(z^S)$ to its hidden state. PG denotes the policy-gradient
token update from outputs after the LM heads; bars are schematic distributions. Routed columns illustrate a math prompt.
One layer pair is drawn: same-family aligns the last three using the identity, cross-family
the last using a shared trainable linear map. Both use crossfade (Figure~\ref{fig:schedule}).}
\label{fig:contrast}
\end{figure}

To answer this question, we introduce \ours{}. It selects late-layer targets
according to the teacher--student relationship and organizes their supervision through \textbf{domain-pure updates}
and a \textbf{per-teacher crossfade} (Figure~\ref{fig:contrast}; Section~\ref{sec:method}).
Each rollout's specialist supplies both token predictions and hidden-state targets.
We align hidden states directly within the same family and use a shared trainable map when widths differ.
The crossfade shifts each specialist's supervision from latent states to token predictions on
its own update clock, adapting the transient latent signal studied by LastOPD
\citep{yang2026lastopd} to domain-routed multi-teacher training. All teachers remain frozen, and the
student needs neither the teachers nor the map at inference.

\vspace{-3pt}
\begin{list}{\contribnum{\arabic{enumi}}}{%
\usecounter{enumi}\setlength{\leftmargin}{1.4em}\setlength{\labelwidth}{0.8em}%
\setlength{\labelsep}{0.6em}\setlength{\topsep}{3pt}\setlength{\itemsep}{3pt}%
\setlength{\parsep}{0pt}\setlength{\partopsep}{0pt}}
\item \begin{minipage}[t]{\linewidth}\textbf{Representation-level multi-teacher OPD.}
We introduce, to our knowledge, the first representation-level multi-teacher OPD method for LLMs. We identify effective layer choices for different teacher panels and show that grouping routed targets by domain stabilizes all-layer representation-only training. \end{minipage}

\item \begin{minipage}[t]{\linewidth}\textbf{Same-family gains beyond individual teachers.}
One $1.5$B student outperforms token-only, representation-only, and uniform-averaging baselines
on all nine benchmarks across math, code, and logic, and surpasses the per-benchmark best teacher
on five of them at the same model size (Figure~\ref{fig:teaser}; Section~\ref{sec:s1}). With the other settings fixed, crossfade improves eight of nine benchmarks over
constant joint weighting.\end{minipage}

\item \begin{minipage}[t]{\linewidth}\textbf{Transfer across teacher panels and initializations.}
With separately developed $7$B Qwen-based teachers, \ours{} outperforms both single-channel
baselines on all six benchmarks (Section~\ref{sec:s2}). Starting from a parameter-merged initialization,
$62$ distillation steps improve performance on five of six benchmarks across all three domains
(Section~\ref{sec:s3}). Further analyses examine layer selection, training dynamics, batch
organization, and generalization (Section~\ref{sec:ablation}).\end{minipage}
\end{list}
Code: \url{https://github.com/fangzy96/Latent-MOPD}.

\section{Related work}
\label{sec:related}

\paragraph{On-policy and representation-level distillation.}
OPD provides teacher supervision on student-generated responses
\citep{agarwal2024onpolicy,gu2024minillm}. Representation matching predates this setting:
FitNets introduced intermediate hints, and subsequent methods aligned hidden states or
attention statistics \citep{romero2015fitnets,sun2019patient,jiao2020tinybert,wang2020minilm}.
OPRD \citep{yang2026oprd} brings hidden-state alignment onto the student's own rollouts,
pairing layers by relative depth. LastOPD \citep{yang2026lastopd}
studies latent collapse across sizes and proposes last-layer supervision that crossfades to
token-only OPD.
PR-OPD \citep{li2026propdprivilegedrepresentationonpolicy} aligns an agent's hidden states with its own
skill-conditioned copy.  These studies focus on supervision from one teacher. We study how to select
and organize representation targets from several specialists.

\paragraph{Multi-teacher distillation.}
Off-policy methods combine teacher outputs or align several teachers' features on fixed
training data \citep{you2017multiteacher,yuan2021reinforced,formont2025taskagnostic}; FuseLLM
combines language models through their output distributions \citep{wan2024fusellm}.
MOPD's reported pipeline \citep{mopd2026} trains domain specialists from a shared SFT checkpoint,
then routes student rollouts to the frozen teachers for token-level distillation. Our experiments
directly reuse existing specialist checkpoints and add hidden-state targets from the same routed
teacher. MT-SDPO verifies teacher candidates against answers \citep{mtsdpo2026}, while Open-MOPD
examines capability imbalance under domain routing \citep{gao2026openmopd}.

\paragraph{Model merging and composition.}
Model soups average compatible model weights \citep{wortsman2022modelsoups}, while task
arithmetic combines task-specific parameter changes \citep{ilharco2023taskarithmetic}.
We also use a parameter merge to initialize the student for $62$ steps of
distillation, testing whether distillation improves over the merged initialization while retaining a single
deployable student. Appendix~\ref{app:related} extends the comparison and discusses
representation diagnostics.

\section{\ours{}: Representation alignment from multiple teachers}
\label{sec:method}

\ours{} distills $N$ frozen teachers $\{\pi_{T_i}\}_{i=1}^N$ into one student $\pi_\theta$.
Each prompt $x$ carries a domain label $d\in\{1,\dots,N\}$ selecting its specialist: math, code
or logic here. The student samples a response, the selected teacher evaluates it,
and the student learns from both that teacher's token predictions and selected hidden states. The same
teacher supplies token targets throughout the response and latent targets at the positions
specified for each regime (Figure~\ref{fig:contrast}).

Let $\hat{y}\sim\pi_\theta(\cdot\mid x)$ denote a student response and $M$ the valid response
positions in a loss calculation. At each prefix $s_t=(x,\hat{y}_{<t})$, $p_{\theta,t}$ and
$p_{T_i,t}$ are next-token distributions over the full vocabulary. We write $\bar p_t$ for the
fixed student scoring distribution and $V_{k,t}$ for its top-$k$ support, with $k=16$.
For a selected layer pair, $z^S_t$ and $z^{T_i}_t$ are the hidden states; the final pair is taken
after final normalization and before the LM heads. We write
$\nu(z)=z/\max(\lVert z\rVert_2,\epsilon)$ for $\ell_2$ normalization and $\sg$ for stop-gradient.
Appendix~\ref{app:prelim} gives the corresponding single-teacher objectives. Representation
matching is used only during training; inference uses the student alone.

\begin{figure}[!t]
\centering
\includegraphics[width=\textwidth]{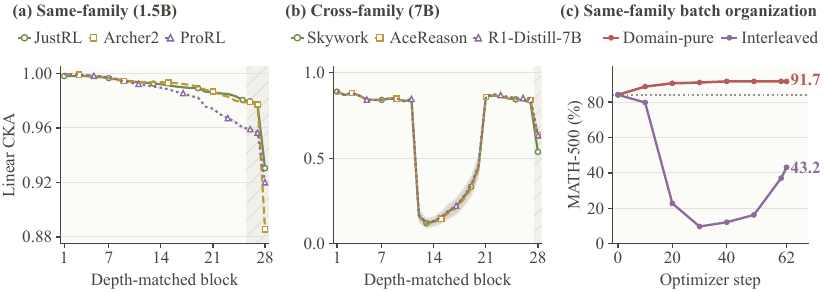}
\caption{\textbf{Representation targets and update organization.}
\textbf{(a,b)} Native linear CKA between matched block outputs on fixed base-student responses
before training. Same-family differences increase near the output (magnified scale);
cross-family similarities show a middle-layer trough. Hatching marks the selected depths;
bands give one prompt-bootstrap standard deviation (Appendix~\ref{app:cka}).
\textbf{(c)} Same-family all-layer representation-only supervision remains stable with
domain-pure updates; interleaved updates collapse and finish far below the base (dotted line).
Both use MATH-500 with $8$ samples per problem (Appendix~\ref{app:detonate}).}
\label{fig:cka}
\end{figure}

\paragraph{Domain-routed token channel (MOPD-style).}
Averaging teacher distributions (Figure~\ref{fig:contrast}a) can dilute a specialist's confident
prediction when the other teachers assign it less probability. MOPD \mbox{\citep{mopd2026}} instead
routes each prompt to its domain's teacher $\pi_{T_d}$. This assignment stays fixed throughout
the response. We use domain-pure batches of $32$ (Figure~\ref{fig:cka}c;
Section~\ref{sec:detonate}). At each response position, we score the student's top-$k$
candidates using teacher-to-student log-probability ratios. Candidate weights renormalize
$\bar p_t$ within $V_{k,t}$, while both log probabilities retain full-vocabulary normalization.
Scaling these rewards by their domain's masked standard deviation gives fixed \mbox{advantages}
$A_{t,v}$ (Appendix~\ref{app:prelim}). The token update is
\begin{equation}
\lopd^{\mathrm{route}} \;=\; -\frac{1}{|M|}\sum_{t\in M}\sum_{v\in V_{k,t}}
\sg[A_{t,v}]\,\log p_{\theta,t}(v).
\label{eq:anch}
\end{equation}
Each position receives one specialist's token targets. The representation channel additionally
supervises hidden states upstream of the head.

\begin{figure}[!t]
\centering
\includegraphics[width=\textwidth]{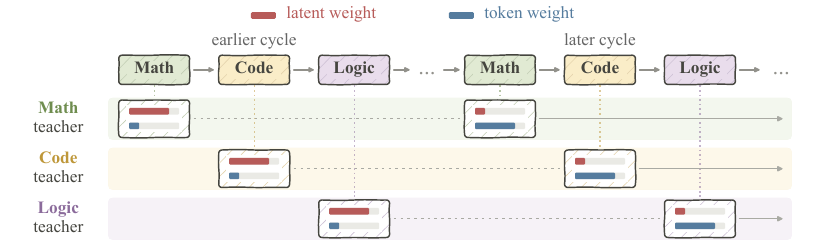}
\caption{\textbf{Crossfade across rotating teachers.} Domain-pure updates rotate through
math, code and logic. Two selected cycles illustrate token weights rising and latent weights
fading on each teacher's own clock (Equation~\ref{eq:cross}); bar lengths are schematic.
Both regimes use this schedule, with different windows.}
\label{fig:schedule}
\end{figure}

\paragraph{Selective late-layer alignment.}
Same-family teachers share the student's architecture and initialization lineage. Their native
CKA remains high through most blocks and decreases near the output (Figure~\ref{fig:cka}a).
We align the final three layers, supported by the layer ablations in Section~\ref{sec:ablation}.
Across scales and training lineages, the middle-layer trough (Figure~\ref{fig:cka}b) cautions
against assuming correspondence at every depth. We align the final state at each selected
token: it integrates prediction-relevant information from preceding computation and directly
feeds the next-token prediction head in each model, giving the target a common functional role.
Within each panel, the teachers
share a hidden width $d_T$. The map $g_\psi:\mathbb{R}^{d_S}\!\to\mathbb{R}^{d_T}$ is the identity in the
same-family setting and, by default, a trainable linear map shared across routed
teachers and selected layers in the cross-family setting. The linear map is initialized by a
ridge fit to domain-routed student--teacher state pairs (Appendix~\ref{app:hp}).
Selecting $3$ of $28$ layers requires $89.3\%$ less selected-state storage
at fixed token count and precision (Appendix~\ref{app:state_storage}).
For $16{,}384$ positions and BF16 width-$1{,}536$ states, the student and teacher tensor copies
total $0.28$\,GiB for three layers, versus $2.63$\,GiB for all $28$.
For one selected layer pair, the weighted penalty at position $t$ is
\begin{equation}
\ell^{\mathrm{route}}_{\mathrm{rep},t} \;=\; \frac{\lambda}{d_T}
\big\lVert \nu\!\left(g_\psi(z^S_t)\right)-\sg\!\left[\nu(z^{T_d}_t)\right]\big\rVert_2^2 .
\label{eq:route}
\end{equation}
Hard domain routing selects only the prompt's specialist: math responses receive math-teacher
states and code responses receive code-teacher states. The loss $\lrep^{\mathrm{route}}$
aggregates these penalties over all response positions
at the last three depth-matched layers in the same-family setting, and over the
final $2{,}000$ response positions at the last layer in the cross-family setting (all positions for shorter responses).
For each selected layer, we average over selected response-token positions across participating
responses, then average equally over layers. Thus selecting three layers does not triple the
representation weight. Gradients update the student and,
cross-family, the shared map; every teacher stays frozen. We use a fixed coefficient
$\lambda$ within each training configuration, with no running loss-share normalization
(reductions in Appendix~\ref{app:hp}).

\paragraph{One schedule: a crossfade on a per-teacher clock.}
Both regimes use a transient representation term, adopting the crossfade form of
LastOPD \citep{yang2026lastopd}. The token channel ramps in as the representation channel fades
out. At global optimizer step $s$, let $u$ count completed updates for the active teacher and
$T_w$ denote its crossfade window:
\begin{equation}
\mathcal{L}^{(s)} \;=\; \alpha(u)\,\lopd^{\mathrm{route}} + \beta(u)\,\lrep^{\mathrm{route}},
\qquad
\alpha(u)=\min\!\Big(1,\tfrac{u}{T_w}\Big),
\qquad
\beta(u)=\max\!\Big(0,\,1-\tfrac{u}{T_w}\Big).
\label{eq:cross}
\end{equation}
Each domain-pure optimizer update uses one pair of weights; only the active teacher's counter
advances afterward. Each teacher receives the same sequence of weights before switching to
token-only updates. We use $T_w=10$ per-teacher steps same-family and $T_w=7$ cross-family. Under the three-teacher
rotation, these windows span $30$ and $21$ of the $62$ global steps, respectively (Figure~\ref{fig:schedule}).
MOPD-style retains a constant token-loss weight throughout training;
representation-only baselines retain the latent objective for the full run without a token term.

\paragraph{Initialization from a parameter merge.}
When the teachers share the student's lineage, their weights can also be
averaged into a zero-training \emph{parameter merge} (a ``model soup'',
\mbox{\citealp{wortsman2022modelsoups}}), combining the experts' weights before distillation begins. We use
this merge to initialize the student for $62$ steps with the same-family crossfade recipe
(Section~\ref{sec:s3}).

\begin{table}[!t]
\centering
\caption{\textbf{Same-family $1.5$B teachers}, distilled into the $1.5$B student for $62$ steps.
Gray rows are teachers; blue rows are \ours{}, with darker blue marking the default last-three-layer setting.
\textbf{Bold} and \underline{underline} mark the best and second-best non-teacher scores;
$\dagger$ exceeds the best teacher.
\textbf{Norm}: base $=0$, best-teacher envelope $=1$.
Appendix~\ref{app:hp} specifies evaluation, Norm and baseline implementation.}
\label{tab:s1}
\tablefont
\setlength{\tabcolsep}{1.4pt}
\begin{tabular}{l ccc ccc ccc @{\hspace{3pt}}|@{\hspace{3pt}} c}
\toprule
& \multicolumn{3}{c}{\textbf{Logic}} & \multicolumn{3}{c}{\textbf{Code}} & \multicolumn{3}{c}{\textbf{Math}} & \\
\cmidrule(lr){2-4}\cmidrule(lr){5-7}\cmidrule(lr){8-10}
Method & GYM & BBH & MuSR & LCB-e & MBPP & MBPP+ & AIME24 & AIME25 & Minerva & Norm \\
\midrule
\rowcolor{basegray}
Teacher \emph{(math)} & 39.1 & 60.5 & 50.3 & 66.8 & 60.4 & 51.1 & 56.5 & 37.7 & 33.6 & -- \\
\rowcolor{basegray}
Teacher \emph{(code)} & 25.6 & 41.6 & 45.1 & 73.5 & 62.0 & 52.6 & 42.1 & 30.0 & 32.2 & -- \\
\rowcolor{basegray}
Teacher \emph{(logic)} & 52.8 & 61.8 & 50.9 & 74.3 & 59.0 & 49.6 & 46.2 & 35.6 & 33.8 & -- \\
Base student & 24.6 & 39.9 & 46.4 & 51.1 & 54.4 & 45.2 & 29.2 & 21.9 & 29.9 & 0.00 \\
Uniform averaging & 40.7 & 60.0 & 50.4 & 72.8 & 61.1 & 50.7 & 48.1 & 34.0 & 34.6$^\dagger$ & 0.79 \\
MOPD-style (token-only) & 51.8 & \underline{65.4}$^\dagger$ & 50.7 & 68.3 & 63.4$^\dagger$ & 54.1$^\dagger$ & 46.0 & 35.0 & 33.5 & 0.90 \\
OPRD-style (all layers) & 38.9 & 57.1 & 50.8 & 68.3 & 59.7 & 49.2 & 43.8 & 32.3 & 32.5 & 0.64 \\
Rep-only (last 1) & 42.7 & 57.2 & 49.1 & 64.6 & 58.7 & 49.9 & 44.6 & 32.3 & 33.1 & 0.64 \\
Rep-only (last 3) & 42.9 & 59.3 & 49.5 & 69.8 & 60.6 & 52.2 & 44.0 & 34.2 & 31.9 & 0.73 \\
\midrule
\rowcolor{ourvariantrow}
\ours{} (constant) & \underline{51.9} & 64.4$^\dagger$ & 50.1 & 71.6 & 63.9$^\dagger$ & 53.6$^\dagger$ & 47.3 & \textbf{37.5} & 33.9$^\dagger$ & 0.94 \\
\rowcolor{ourvariantrow}
\ours{} (OPRD-style) & 51.5 & 65.2$^\dagger$ & 49.5 & \textbf{73.9} & 62.7$^\dagger$ & 52.5 & 47.9 & 36.3 & 33.9$^\dagger$ & 0.94 \\
\rowcolor{ourvariantrow}
\ours{} (last 1) & 51.3 & 64.4$^\dagger$ & \underline{51.5}$^\dagger$ & 71.6 & \underline{65.5}$^\dagger$ & \underline{54.6}$^\dagger$ & \textbf{51.0} & 35.8 & \underline{34.8}$^\dagger$ & \underline{0.99} \\
\rowcolor{ourrow}
\textbf{\ours{} (last 3)} & \textbf{52.4} & \textbf{66.3$^\dagger$} & \textbf{52.4$^\dagger$} & \underline{73.1} & \textbf{65.7$^\dagger$} & \textbf{55.7$^\dagger$} & \underline{50.8} & \underline{36.9} & \textbf{36.3$^\dagger$} & \textbf{1.05} \\
\bottomrule
\end{tabular}
\end{table}

\section{Experiments}
\label{sec:exp}

\subsection{Setup}
\label{sec:setup}

\paragraph{Models, regimes and protocol.}
We distill existing specialist checkpoints into DeepSeek-R1-Distill-Qwen-$1.5$B
\citep{deepseekr1}, without additional teacher training.
Our main \textbf{same-family} regime uses three RL-tuned $1.5$B teachers
from the same model lineage as the student, one per domain.
\textbf{Cross-family} uses three separately developed $7$B teachers with hidden width
$3{,}584$ (student: $1{,}536$). Here
``cross-family'' denotes different scale and post-training; all four models have $28$ blocks
and use a Qwen backbone. \emph{Merge initialization} starts the student from the same-family
teachers' parameter merge (Table~\ref{tab:s3}). Training prompts come from DAPO-Math-17k,
OpenCodeReasoning and Reasoning Gym \citep{yu2026dapo,ahmad2025opencodereasoning,stojanovski2025reasoninggym}. The pool contains $17{,}856$ prompts
($5{,}952$ per domain).
Training and evaluation share no problem instances, as checked before training.
Domain-pure blocks of $32$ rotate without shuffling: $62$ updates process $1{,}984$
prompt presentations and generate $7{,}936$ responses ($4$ per prompt), without verifier rewards.
The main tables evaluate each trained student at its final checkpoint (step $62$).

\vspace{-3pt}
\paragraph{Evaluation and baselines.}
The main tables cover \emph{math}, \emph{code} and \emph{logic}; Figure~\ref{fig:curves}b tests
general ability on suites excluded from distillation. GYM is a frozen $925$-problem set
with a $16$k-token budget (scoring and datasets in Appendices~\ref{app:hp} and~\ref{app:datasets}).
Tables~\ref{tab:s1}--\ref{tab:s2} compare the base student, teachers, \emph{MOPD-style (token-only)}
and representation-only baselines. The MOPD-style baseline retains domain-based teacher routing
\citep{mopd2026}. Its token estimator matches \ours{}: student-selected top-$k$ support with
full-vocabulary log-probability ratios and student-normalized candidate weights. \emph{OPRD-style (all layers)} adapts
\citet{yang2026oprd} to routed multi-teacher supervision of all $28$ layers;
\emph{Rep-only (last 1)} and \emph{(last 3)} use the final one or three. These baselines retain representation
supervision throughout training without a token loss. \emph{\ours{} (OPRD-style)} adds the
routed token channel to all-layer supervision, with both weights constant.
The routed methods share prompts, batches and optimizer, with the reported differences in
teachers, initialization, objective and schedule. Appendix~\ref{app:hp} distinguishes our MOPD-style
implementation from the published system and specifies the \emph{uniform averaging}
protocol. Table~\ref{tab:s3} compares a zero-training \emph{parameter merge} with students
distilled from it. Norm measures aggregate improvement over the base divided by the summed
base-to-best-teacher gaps: base $=0$ and best-teacher envelope $=1$.

\subsection{Same-family teachers: surpassing individual specialists}
\label{sec:s1}

With three existing same-family $1.5$B specialists, the default last-three-layer student
surpasses the per-benchmark best teacher on five of nine benchmarks: BBH, MuSR,
MBPP, MBPP+, and Minerva (Table~\ref{tab:s1}). These gains span logic, code and math in one
student with the same parameter count as each teacher. Norm reaches $1.05$, above the
best-teacher envelope at $1$, compared
with $0.90$ for MOPD-style, $0.64$--$0.73$ for representation-only baselines, and $0.94$ for
\ours{} (OPRD-style).

\ours{} also improves over the token-only, representation-only and averaging baselines on
all nine benchmarks. Relative to MOPD-style, Minerva rises from $33.5$ to $36.3$, LiveCodeBench-easy
from $68.3$ to $73.1$, and MuSR from $50.7$ to $52.4$. These gains use the same prompts and
number of updates as the routed single-channel baselines. Among joint methods, crossfade improves
eight of nine benchmarks over the matched constant-weight, last-three-layer control
(Norm $1.05$ vs.\ $0.94$). Our complete default configuration also outperforms
\ours{} (OPRD-style) on eight of nine benchmarks.

\subsection{Extension to cross-family teachers}
\label{sec:s2}

\begin{table}[!t]
\centering
\caption{\textbf{Cross-family $7$B teachers.} Skywork-OR1-Math,
AceReason-Nemotron and R1-Distill-7B, each $\sim\!2.3\times$ the student's hidden width,
distilled into the same $1.5$B student. Darker blue highlights the MLP variant with the highest Norm.
Score markings and Norm follow Table~\ref{tab:s1}.}
\label{tab:s2}
\tablefont
\setlength{\tabcolsep}{3.0pt}
\begin{tabular}{l cc cc cc | c}
\toprule
& \multicolumn{2}{c}{\textbf{Logic}} & \multicolumn{2}{c}{\textbf{Code}} & \multicolumn{2}{c}{\textbf{Math}} & \\
\cmidrule(lr){2-3}\cmidrule(lr){4-5}\cmidrule(lr){6-7}
Method & GYM & BBH & MBPP & MBPP+ & AIME24 & AIME25 & Norm \\
\midrule
\rowcolor{basegray}
Teacher \emph{(math)} & 46.1 & 54.9 & 78.2 & 64.3 & 61.9 & 52.9 & -- \\
\rowcolor{basegray}
Teacher \emph{(code)} & 47.6 & 74.3 & 83.7 & 70.2 & 64.4 & 51.0 & -- \\
\rowcolor{basegray}
Teacher \emph{(logic)} & 42.0 & 64.2 & 80.2 & 67.2 & 55.2 & 41.5 & -- \\
Base student & 24.6 & 39.9 & 54.4 & 45.2 & 29.2 & 21.9 & 0.00 \\
MOPD-style (token-only) & 29.2 & 45.8 & 55.4 & 46.8 & 36.5 & 30.2 & 0.16 \\
Rep-only (last 1) & 23.6 & 35.3 & 53.2 & 46.2 & 25.6 & 22.7 & $-0.05$ \\
\midrule
\rowcolor{ourvariantrow}
\ours{} (last 3, Linear) & \textbf{34.6} & \underline{47.6} & \textbf{59.4} & \textbf{50.9} & 36.9 & 30.6 & 0.25 \\
\rowcolor{ourvariantrow}
\ours{} (last 1, Linear) & \underline{34.1} & \textbf{49.5} & 57.9 & 48.5 & \underline{39.4} & \textbf{31.7} & \underline{0.26} \\
\rowcolor{ourrow}
\textbf{\ours{} (last 1, MLP)} & 32.5 & 47.0 & \underline{58.9} & \underline{50.4} & \textbf{42.9} & \underline{31.0} & \textbf{0.27} \\
\bottomrule
\end{tabular}
\end{table}

To test transfer beyond the same-family setting, we use three separately developed $7$B teachers.
The default last-layer \ours{} with a linear map outperforms
token-only and representation-only baselines on all six benchmarks across math, code and logic
(Table~\ref{tab:s2}). Relative to MOPD-style, AIME24 improves from $36.5$ to $39.4$ and MBPP from $55.4$
to $57.9$. Logic sees the largest gains: GYM rises from $29.2$ to $34.1$ and BBH from $45.8$ to
$49.5$. The Norm score rises from $0.16$ for MOPD-style to $0.26$ for \ours{}.
A two-layer MLP also outperforms both single-channel baselines on all six benchmarks
(Norm $0.27$); the two projector configurations split benchmark wins three to three.

\subsection{Exploring a parameter-merged initialization}
\label{sec:s3}

We also explore distillation from a parameter-merged initialization (Table~\ref{tab:s3}).
After $62$ steps, \ours{} improves five of six benchmarks across math, code and logic.
GYM rises from $46.3$ to $52.2$, BBH from $60.6$ to $63.4$, and AIME24 from $48.1$ to $53.8$.
Norm increases from $0.87$ to $1.03$, above token-only ($0.97$) and representation-only
($0.91$) training from the same merge. The student also exceeds the best same-family teacher
on BBH, MBPP and MBPP+.

\begin{table}[!t]
\centering
\caption{\textbf{Parameter-merge initialization.} Uniform merging and $62$-step
distillation on the same six benchmarks as Table~\ref{tab:s2}.
Gray rows are teachers; blue marks \ours{}. Bold and underline mark the best and second-best
non-teacher scores, respectively.}
\label{tab:s3}
\tablefont
\setlength{\tabcolsep}{3.0pt}
\begin{tabular}{l cc cc cc | c}
\toprule
& \multicolumn{2}{c}{\textbf{Logic}} & \multicolumn{2}{c}{\textbf{Code}} & \multicolumn{2}{c}{\textbf{Math}} & \\
\cmidrule(lr){2-3}\cmidrule(lr){4-5}\cmidrule(lr){6-7}
Method & GYM & BBH & MBPP & MBPP+ & AIME24 & AIME25 & Norm \\
\midrule
\rowcolor{basegray}
Teacher \emph{(math)} & 39.1 & 60.5 & 60.4 & 51.1 & 56.5 & 37.7 & -- \\
\rowcolor{basegray}
Teacher \emph{(code)} & 25.6 & 41.6 & 62.0 & 52.6 & 42.1 & 30.0 & -- \\
\rowcolor{basegray}
Teacher \emph{(logic)} & 52.8 & 61.8 & 59.0 & 49.6 & 46.2 & 35.6 & -- \\
Base student & 24.6 & 39.9 & 54.4 & 45.2 & 29.2 & 21.9 & 0.00 \\
Parameter merge \emph{(start)} & 46.3 & 60.6 & \underline{64.9} & \textbf{55.4} & 48.1 & 34.0 & 0.87 \\
\;+MOPD-style (token-only) & \textbf{52.3} & \underline{62.2} & 63.8 & 53.8 & 50.6 & \textbf{37.3} & \underline{0.97} \\
\;+Rep-only (last 3) & 46.7 & 61.2 & 62.0 & 52.7 & \textbf{54.6} & 36.5 & 0.91 \\
\rowcolor{ourrow}
\;+\ours{} & \underline{52.2} & \textbf{63.4} & \textbf{65.6} & \underline{54.9} & \underline{53.8} & \underline{36.7} & \textbf{1.03} \\
\bottomrule
\end{tabular}
\end{table}

\begin{figure}[!t]
\centering
\includegraphics[width=\textwidth]{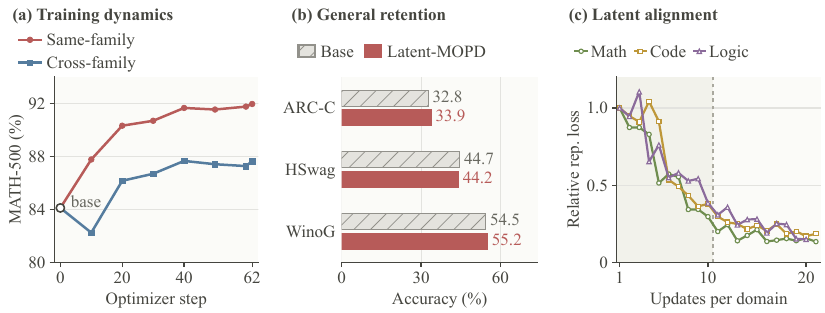}
\caption{\textbf{Training dynamics, general retention and representation alignment.}
\textbf{(a)} Default same-family (last-three-layer) and cross-family (last-layer) \ours{}
on MATH-500, with $8$ samples per problem.
\textbf{(b)} Same-family general retention under $0$-shot likelihood scoring.
\textbf{(c)} Same-family raw representation loss, normalized to each domain's first logged
value and excluding $\beta$. Crossfade lasts $10$ updates per domain (shaded), followed by
token-only supervision. Appendix~\ref{app:hp} specifies scoring and loss measurement.}
\label{fig:curves}
\end{figure}

\section{Ablations and analysis}
\label{sec:ablation}

\begin{figure}[!t]
\centering
\includegraphics[width=\textwidth]{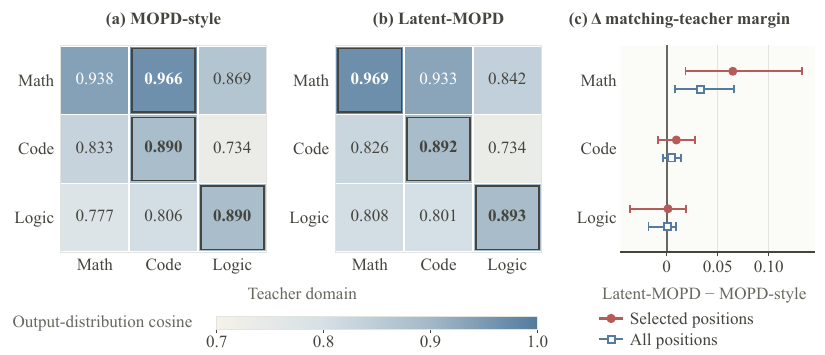}
\caption{\textbf{Domain-conditioned teacher matching after same-family distillation.}
\textbf{(a,b)} Mean next-token probability cosine on teacher-disagreement positions:
input domains are rows and teachers are columns, on one shared color scale; outlines mark row maxima.
\textbf{(c)} Change in matching-teacher margin (\ours{} minus MOPD-style) on selected and all
positions, with paired $95\%$ instance-bootstrap intervals. Appendix~\ref{app:output_alignment}
defines the selection rule, margin and measurement protocol.}
\label{fig:output_alignment}
\end{figure}

Tables~\ref{tab:s1}--\ref{tab:s2} compare token-only, representation-only and joint supervision;
Table~\ref{tab:s1} also tests crossfade against constant weights. We examine layer selection,
update organization, training dynamics and teacher matching (Figures~\ref{fig:cka},
\ref{fig:curves} and~\ref{fig:output_alignment}); Appendix~\ref{app:cases} compares solutions.
For the last-layer and last-three-layer \ours{} comparisons, coefficient, response positions
and crossfade are fixed within each regime.

\paragraph{Layer selection depends on the teacher panel.}
In the same-family setting, three-layer supervision improves eight of nine benchmarks,
raising Norm from $0.99$ to $1.05$ (Table~\ref{tab:s1}).
The representation-only controls also favor three layers: they outperform the last-layer
and all-layer alternatives on seven of nine benchmarks each (Norm $0.73$ vs.\ $0.64$).
Cross-family with linear projectors, last-layer supervision reaches Norm $0.26$, versus $0.25$ for three layers.
Benchmark wins split three to three; one layer retains the aggregate gain with fewer representation
targets. These comparisons test the layer choices motivated by native CKA (Figure~\ref{fig:cka}a,b);
Appendices~\ref{app:obs}--\ref{app:cka} give further comparisons and masking diagnostics.

\vspace{-2pt}
\paragraph{Training dynamics across teacher panels.}
Figure~\ref{fig:curves}a tracks each regime's default configuration on $500$ MATH problems
($8$ samples each; Appendix~\ref{app:hp}). From the base score of $84.1$, the same-family run
reaches $90.3$ at step $20$ and $92.0$ at step $62$; cross-family reaches $86.2$ and $87.7$,
respectively. Both remain above the base from step $20$ through the later token-only phase.

\vspace{-2pt}
\paragraph{Domain-pure updates stabilize representation-only training.}
\label{sec:detonate}
Routing assigns a teacher to each prompt, but does not determine whether an optimizer step
contains one domain or several. We examine this choice in a same-family all-layer representation-only
control, holding teachers, loss, coefficient and budget fixed (Figure~\ref{fig:cka}c).
Domain-pure blocks of $32$ complete all $62$ steps, reaching $91.7$ on in-loop MATH-500
validation. Interleaving the same rows collapses MATH-500 to $22.8$ at step $20$ and $9.8$
at step $30$; it remains far below the base at step $62$ ($43.2$ vs.\ $84.1$).
These results show that domain-pure updates stabilize all-layer representation-only training
under the tested configuration (Appendix~\ref{app:detonate}).

\paragraph{General-benchmark performance stays close to the base.}
On three suites absent from the training pool, the same-family student's scores remain within
$1.1$ points of the base (Figure~\ref{fig:curves}b): ARC-Challenge and WinoGrande improve,
while HellaSwag changes from $44.7$ to $44.2$. These tests score answer-option likelihoods
without sampling (Appendix~\ref{app:hp}).

\paragraph{Learning representation targets from three specialists.}
Figure~\ref{fig:curves}c tracks the same-family student's raw representation loss on its
training rollouts, normalized within each domain. At update $10$ per domain, the math,
code and logic losses fall to $0.30$, $0.38$ and $0.39$ of their first logged values.
All three remain low after representation supervision ends, reaching $0.13$, $0.19$ and
$0.15$ at their final recorded updates. These losses exclude $\beta$, so their decline is
not simply the prescribed weight decay. The student reduces its mismatch with all three
specialists without maintaining the latent objective throughout training
(measurement details in Appendix~\ref{app:alignment_dynamics}).

\paragraph{Matching the routed specialist after training.}
Figure~\ref{fig:output_alignment} compares the final same-family students with all three
teachers on identical prefixes. On teacher-disagreement positions, the math row's maximum
shifts from the code teacher under MOPD-style to the math teacher under \ours{}; code and logic
retain their domain-matched maxima. Matching-teacher margin measures mean similarity to
the domain specialist minus the closest alternative. The math margin increases by $0.0648$
on selected positions and $0.0327$ over all positions, with positive paired intervals in both
cases (protocol in Appendix~\ref{app:output_alignment}).

\paragraph{Case studies.}
Appendix~\ref{app:cases} compares MOPD-style and \ours{} solutions with the routed specialist's
reference. In these examples, \ours{} preserves the recurrence boundary, rewrite invariant,
and counting constraints that MOPD-style misses. These differences motivate supervising hidden
states, which can encode intermediate reasoning information \citep{gurnee2026workspace}.

\section{Conclusion}
\label{sec:conclusion}

\ours{} integrates existing specialists through routed late-layer alignment, shared width
projection, domain-pure updates and per-teacher crossfades. The same-family $1.5$B student
outperforms token-only and representation-only baselines on all nine benchmarks across math,
code and logic, and surpasses the per-benchmark best teacher on five. With cross-family teachers,
it outperforms both single-channel baselines on all six benchmarks. From a parameter-merged
initialization, it improves five of six benchmarks over the initial merge. Teachers remain frozen;
deployment uses only the student, with no projection map. Appendix~\ref{app:limits} discusses scope and extensions.

\bibliography{refs}
\bibliographystyle{preprint}

\appendix
\section{Experimental setup and reproduction}
\label{app:hp}

\paragraph{Models, teachers, and routing.}
The student is DeepSeek-R1-Distill-Qwen-$1.5$B \citep{deepseekr1}. The same-family panel is
JustRL-DeepSeek-$1.5$B (math), Archer2-Code-$1.5$B (code), and Nemotron-ProRL-$1.5$B (logic),
all RL-tuned from the student's lineage.\footnote{\raggedright Public same-family checkpoints (ProRL at revision \texttt{v2}):
\url{https://huggingface.co/hbx/JustRL-DeepSeek-1.5B};
\url{https://huggingface.co/Fate-Zero/Archer2.0-Code-1.5B-Preview};
\href{https://huggingface.co/nvidia/Nemotron-Research-Reasoning-Qwen-1.5B/tree/v2}{\nolinkurl{https://huggingface.co/nvidia/Nemotron-Research-Reasoning-Qwen-1.5B}}.} The cross-family panel is Skywork-OR1-Math-$7$B (math),
AceReason-Nemotron-$7$B (code), and DeepSeek-R1-Distill-Qwen-$7$B (logic).\footnote{\raggedright Public cross-family checkpoints (Skywork-OR1-Math-7B and AceReason-Nemotron-7B are RL-tuned from DeepSeek-R1-Distill-Qwen-7B):
\url{https://huggingface.co/Skywork/Skywork-OR1-Math-7B};
\url{https://huggingface.co/nvidia/AceReason-Nemotron-7B};
\url{https://huggingface.co/deepseek-ai/DeepSeek-R1-Distill-Qwen-7B}.} All teachers are existing
public checkpoints, reused without additional teacher training for this study. Each training prompt
carries a domain label in $\{$math, code, logic$\}$. Both channels are routed by that label to the
\emph{same} per-domain teacher: in the same-family panel, math\,$\to$\,JustRL,
code\,$\to$\,Archer2, and logic\,$\to$\,ProRL. The
domain-routed token channel (MOPD-style, Equation~\ref{eq:anch}) scores the tokens of each student response against the teacher
for its prompt's domain, in domain-pure batches; the domain-routed representation channel
(Equation~\ref{eq:route}) uses that teacher's selected hidden states as its target.
The parameter merge of Section~\ref{sec:method} is the uniform weight average
of the three same-family teachers.

\paragraph{Optimization and rollout protocol.}
Every optimizer step samples $4$ responses at temperature $1.0$ for each of $32$ prompts and
applies a constant learning rate of $10^{-5}$; the $62$ steps generate $7{,}936$ responses,
matching the on-policy budget of LastOPD \citep{yang2026lastopd}.
We use AdamW with betas $(0.9,0.999)$, weight decay $0.01$, and $\epsilon=10^{-8}$.
The routed token-only, representation-only and combined methods share prompts, batches and
optimizer, with the teacher panel, initialization, objective and schedule specified for each arm.
Uniform averaging follows the same protocol. Rollouts and evaluation
apply the model's chat template with \texttt{enable\_thinking=False}; during distillation,
teachers score the resulting student responses. Training responses are capped at
$16{,}384$ tokens, with repetition penalty $1.0$. The token update uses the student's top-$16$
candidates and the fixed-advantage surrogate in Equation~\ref{eq:anch}.

\paragraph{Prompt pool and update counts.}
The pool contains $17{,}856$ prompts: $5{,}952$ each for math, code and logic. We disable data shuffling and
use domain-pure blocks of $32$ in math--code--logic order. The $62$-step budget therefore contains
$21$, $21$ and $20$ updates from the three domains, respectively: $672$, $672$ and $640$ prompt
presentations, for $1{,}984$ in total. Four responses per prompt give $128$ generated responses
per update and $7{,}936$ over the run. These are counts of prompt presentations and generated responses; the pool size
does not denote the number of prompts processed within the $62$-step budget.

\paragraph{Token-loss reduction and accumulation.}
The token channel uses the single-epoch policy-gradient surrogate in Equation~\ref{eq:anch}.
For each microbatch, it sums over the $16$ candidate tokens at each position, then takes a
masked mean over valid response positions across its responses; the denominator is the valid
position count plus $10^{-8}$. Padding contributes neither loss nor count. Responses therefore
contribute in proportion to their valid lengths within this calculation. Domain-wise advantage
scales are computed on the rollout batch before microbatch packing (Appendix~\ref{app:prelim}).
On the eight-worker training setup, each worker has a local optimizer minibatch of $16$
responses. A dynamically packed microbatch with $n_c$ responses contributes $n_c/16$ times its
loss to gradient accumulation. This factor multiplies the complete microbatch loss after
applying the configured channel weights. The actor makes one optimizer update after
accumulating the local minibatch, with one optimization epoch per rollout batch.

\paragraph{Position and layer normalization.}
\label{app:rep_reduction}
For response $b$, let $M_b$ be its valid response-token positions and $S_b\subseteq M_b$ the
positions selected for representation supervision. In \ours{}, $S_b=M_b$ same-family; cross-family,
$S_b$ contains the last $\min(2000,|M_b|)$ valid response positions. Padding and unselected
positions do not enter the denominator. Let $\mathcal{C}$ denote the responses participating
in one masked-mean loss calculation and $\mathcal{A}$ the selected layer pairs. Writing
$\ell^{(\ell)}_{\mathrm{rep},b,t}$ for the routed penalty in Equation~\ref{eq:route} at layer
pair $\ell$, the position and layer reduction is
\begin{equation}
R_{\mathcal{C}} = \frac{1}{|\mathcal{A}|}
\sum_{\ell\in\mathcal{A}}
\frac{\displaystyle\sum_{b\in\mathcal{C}}\sum_{t\in S_b}
\ell^{(\ell)}_{\mathrm{rep},b,t}}
{\displaystyle\sum_{b\in\mathcal{C}}|S_b|}.
\label{eq:rep_reduction}
\end{equation}
The selected layer pairs are averaged, so using three layers does not multiply the loss
coefficient by three. Within this calculation, each selected token has equal weight; responses
contribute in proportion to their selected-token counts. This formula describes the reduction
inside the representation-loss calculation, separately from accumulation across optimizer
microbatches and devices. The token objective uses all valid response positions in both regimes.

\paragraph{Representation targets and crossfade.}
The routed representation penalty is normalized by feature norm and hidden width
(Equation~\ref{eq:route}). For the crossfade variants, we use fixed coefficients $\lambda=1$ same-family and $\lambda=2000$
cross-family, with no running loss-share normalization. The crossfade of
Equation~\ref{eq:cross} multiplies the representation term by $\beta(u)$ and the token term by $\alpha(u)$, with
$T_w=10$ \emph{per-teacher} steps for the same-family panel and $T_w=7$ for the cross-family one
(the three domains rotate per block, so the fades span $30$ and $21$ global optimizer steps). The
default same-family configuration applies the representation term at each of the last three layers
and all response positions. The cross-family default uses the last layer and the final $2{,}000$
response positions, or all positions when the response is shorter. Each regime also includes a
layer-selection variant with the same coefficient, position scope and crossfade window:
last-layer supervision same-family, and last-three-layer supervision cross-family. Layer losses
are averaged as above. The respective per-worker packing budgets are $24{,}576$ and $18{,}432$ tokens.

\paragraph{Teacher-local clock.}
The counter $u$ in Equation~\ref{eq:cross} is the number of updates already completed for the
active teacher, so its first update uses $u=0$. Each optimizer update uses a fixed pair
$(\alpha(u),\beta(u))$ across all its responses and token positions; only the active teacher's
counter advances after that update. If $s=1,\ldots,62$ is the global step, the
math--code--logic rotation gives $u=\lfloor(s-1)/3\rfloor$ for the active teacher. Thus the
first three global steps use $\alpha=0$ and $\beta=1$. Same-family, the representation term
remains active through step $30$ and is zero from step $31$ onward; cross-family, it remains
active through step $21$ and is zero from step $22$ onward. All three teachers receive the same
sequence of representation weights during their respective crossfade windows.

\paragraph{Constant-weight schedule control.}
The \emph{\ours{} (constant)} row in Table~\ref{tab:s1} matches the default same-family
last-three-layer configuration except that $\alpha(u)=\beta(u)=1$ for all $62$ updates.
It uses $\lambda=1$, all valid response positions, the identity map and a $24{,}576$-token
per-worker packing budget, with the same teachers, prompt pool, domain-pure batches and optimizer.
This comparison tests the paired channel schedule against sustained joint supervision.

\paragraph{Representation-alignment dynamics.}
\label{app:alignment_dynamics}
Figure~\ref{fig:curves}c uses the default same-family last-three-layer run. At each
domain-pure optimizer step, the logged raw representation loss belongs to the active
domain's teacher and excludes both the coefficient $\lambda$ and the schedule weight
$\beta$; here $\lambda=1$. Let $r_{d,j}$ denote this logged loss on the $j$th update
for domain $d$. We plot $r_{d,j}/r_{d,1}$, with separate denominators for math, code
and logic: $3.90603\times10^{-5}$, $3.465517\times10^{-6}$ and
$1.770245\times10^{-5}$, respectively. These first observations occur at global steps
$1$, $2$ and $3$; no pre-update (step-$0$) value is prepended.
The $62$ global steps supply $21$, $21$ and $20$ updates for the three domains.
All recorded points are shown without smoothing. The first $10$ updates of each
domain use crossfade, and $\beta=0$ from its $11$th update onward; teachers remain frozen.
We retain the recorded \texttt{actor/rep\_loss} scalar and its logging reduction across
microbatch and worker records, rather than recomputing a global token-pooled loss.
These are measurements on evolving on-policy rollouts, rather than a fixed validation
input set. Their normalized values describe within-domain training trajectories and
do not compare absolute representation-error scales between teachers.

\paragraph{Merge-initialized training.}
All three distilled students in Table~\ref{tab:s3} start from the uniform average of the
same-family teachers and train for $62$ domain-pure updates. \ours{} uses $\lambda=1$,
the last three layers, all valid response positions and a $10$-step per-teacher crossfade
on the three-domain cycle. Its per-worker packing budget is $18{,}432$ tokens for policy
updates and log-probability scoring; each generated response is capped at $16{,}384$ tokens.
Representation-only supervises the final three layers with $\lambda=2000$ throughout training.

\paragraph{Projector construction.}
Same-family runs use the identity map. The cross-family default uses a bias-free linear map
$g_\psi:\mathbb{R}^{1536}\!\to\!\mathbb{R}^{3584}$.
Each worker maintains one map shared across its routed teacher targets and selected layer pairs.
The map is initialized from a closed-form ridge fit: $42$ student rollouts provide $134{,}843$
domain-routed pairs of student and teacher last-layer states at response positions. A random
$90\%$ split supplies $S$ and $T$ for
$W=(S^\top S+\lambda_r I)^{-1}S^\top T$, with $\lambda_r=10^{-3}d_S$.
The map is optimized jointly with the student and stored per worker. The default cross-family
run reloads the ridge initialization at process restarts. The map is discarded after training;
inference uses only the student.

The last-layer MLP variant in Table~\ref{tab:s2} replaces the linear map with two bias-free
linear layers, $1536\!\to\!6144\!\to\!3584$, with GELU between them. It uses the same
per-worker map sharing across routed targets and is randomly initialized, whereas the default
linear map uses the ridge initialization. The MLP has $31{,}457{,}280$ parameters versus
$5{,}505{,}024$ for the linear map ($5.71\times$). Other settings are unchanged; the comparison changes both projector architecture and
initialization.

\paragraph{Storage of selected representations.}
\label{app:state_storage}
For one routed active teacher, storing one copy each of the selected student and teacher states
requires $M=LN(d_S b_S+d_T b_T)$ bytes, where $L$ is the number of selected layer pairs,
$N$ is the number of concurrently represented response positions, and $b_S,b_T$ are bytes per
element. Holding the other quantities fixed, selecting three rather than $28$ layers reduces
this storage by $1-3/28\approx89.3\%$. For example, with $N=16{,}384$,
$d_S=d_T=1{,}536$ and BF16 storage ($b_S=b_T=2$), the count is about $2.63$\,GiB for all $28$ layers
and $0.28$\,GiB for the last three; FP32 storage doubles both values. These are analytic
feature-storage counts, not measured peak GPU memory. They exclude model weights, optimizer
state, attention activations, normalization and projection buffers, and other autograd storage.
The student still backpropagates through the full backbone. Materializing all hidden states
before selecting layers can also retain the full transient allocation, so the storage ratio
does not imply the same reduction in training peak memory.

\paragraph{Evaluation and checkpoint selection.}
Main benchmark scores are
measured at the final checkpoint, by default at temperature $0.7$ and top-$p$ $0.95$; math and reasoning suites
use sampled means (AIME24/25 with $16$ samples per problem; Minerva with $4$;
BBH, MuSR and Reasoning Gym with $1$, relying on problem count;
LiveCodeBench with $4$ at temperature $0.6$), and the general suites use
answer-option accuracy, using length-normalized log likelihoods for ARC-Challenge and HellaSwag
and unnormalized log likelihoods for WinoGrande. All AIME25 evaluations use temperature
$0.6$, $16$ samples per problem,
top-$p$ $0.95$ and a $31{,}744$-token generation cap.
AIME24 and Minerva contain $30$ and $272$ problems, respectively, and both use a
$31{,}744$-token generation cap.
We never select the best intermediate checkpoint.

\paragraph{In-loop validation.}
MATH-500 and GYM-200 validation use $500$ and $200$ fixed problems, respectively, with eight
sampled responses per problem, temperature $0.7$, top-$p$ $0.95$ and a $16{,}384$-token generation cap. Scores average
correctness over responses and problems.

\paragraph{Offline evaluation harness.}
For MBPP and MBPP+, we generate $2$ programs for each of $378$ tasks, with a
$16{,}384$-token cap, and score the same programs using EvalPlus~0.3.1.\footnote{\url{https://github.com/evalplus/evalplus}} We report
pass@1, averaging correctness over samples and tasks. MBPP requires passing the base
tests; MBPP+ requires passing both the base and additional tests. Failed extraction
counts as incorrect.
For LiveCodeBench, generations are produced for all
$165$ easy problems dated on or after 2024-01-01, and pass@1 is scored on the $67$ easy problems
dated on or after 2024-08-01 with the official test suites ($4$ samples per
problem, temperature $0.6$, top-$p$ $0.95$; LiveCodeBench uses a $16{,}384$-token budget because these
long-chain models truncate heavily on its problems at $8{,}192$ tokens). BBH uses all $27$ task configurations with the first $40$ problems of each
($1{,}080$ problems, identical for every model), one sample per problem at temperature $0.7$ with
an $8{,}192$-token budget, answers extracted from \verb|\boxed{}| with normalized matching.
MuSR uses $750$ problems spanning its three subtasks (murder mysteries, object placements and
team allocation), with one sample per problem and an $8{,}192$-token budget.
GYM uses our frozen Reasoning Gym evaluation set ($38$ tasks $\times\,25$ problems,
generator seed $4242$, \texttt{reasoning-gym}~0.1.19) under a $16{,}384$-token budget, scored by each task's own verifier. A byte-level check against the
frozen questions excludes one task, leaving $925$ scored problems ($37$ tasks). Generators are frozen to files because seeds do not reproduce across registry versions.
Environment pins: \texttt{torch}~2.8.0, \texttt{vllm}~0.11.0, \texttt{transformers}~4.57.3.

\paragraph{Baseline implementation.}
Our MOPD-style baseline retains MOPD's domain routing and uses the same token estimator,
prompts, batches, optimizer and budget as \ours{}. It is not a replication of the published system
\citep{mopd2026}, which trains same-origin teachers at a much larger scale on a different domain
mix and uses a policy-gradient reverse-KL estimator by default. The published system's top-$k$ variant uses
teacher-selected support and a correction for truncation. Our token channel uses student-selected
support, full-vocabulary log-probability ratios and student-normalized candidate weights in a
fixed-advantage policy-gradient update (Appendix~\ref{app:prelim}). Our Norm also uses the headroom-weighted definition below, not MOPD's uniform average of per-domain normalized scores.

Uniform averaging replaces the routed specialist's token distribution with the equally
weighted mean of all three teachers' probabilities on each student-generated response.
It otherwise uses the same training and evaluation protocol as MOPD-style: teacher checkpoints,
student initialization, prompt pool, domain-pure batch order, optimizer, update budget,
token estimator, response limits, numerical precision and microbatch packing.
Both token-only baselines retain a constant token-loss weight and are evaluated at the final
checkpoint. Their comparison isolates how the teacher target is formed.

The same-family \emph{OPRD-style (all layers)} baseline adapts representation distillation
\citep{yang2026oprd} to domain-routed teachers, supervising all $28$ layers; \emph{Rep-only
(last 3)} and \emph{Rep-only (last 1)} use the final three and one, respectively.
All three use the final $\min(2000,|M_b|)$ valid positions
per response, fixed $\lambda=2000$ for $62$ steps, and an $18{,}432$-token per-worker
microbatch packing budget.
Cross-family, \emph{Rep-only (last 1)} matches default \ours{} in its ridge-initialized,
bias-free linear projector, per-worker map sharing, joint student--projector optimization,
last-layer supervision, final $\min(2000,|M_b|)$ valid response positions, $\lambda=2000$,
$18{,}432$-token per-worker packing budget and remaining training settings.
All representation-only baselines retain
the representation objective throughout all $62$ updates, without token loss or crossfade.
The same-family \emph{\ours{} (OPRD-style)} variant combines our routed token channel with all $28$
layers, the last $2{,}000$ valid positions, $\lambda=2000$ and an $18{,}432$-token packing budget.
Both channel weights remain constant for all $62$ steps. Its coefficient, layers, positions, schedule and packing budget thus differ from the default.

\paragraph{Headroom-weighted normalization.}
Within each of Tables~\ref{tab:s1}--\ref{tab:s3}, let $b_j$ be the base-student score and $u_j$
the best individual teacher's score on benchmark $j$. We report
\begin{equation}
\mathrm{Norm}(x)=\frac{\sum_j(x_j-b_j)}{\sum_j(u_j-b_j)}.
\end{equation}
The sums run over the benchmarks reported in that table: the base student scores $0$
and the per-benchmark best-teacher envelope scores $1$. This weights each benchmark's
fraction of closed headroom by its available headroom, avoiding disproportionate influence from
benchmarks where the teacher is only slightly above the base.

\section{Datasets and evaluation coverage}
\label{app:datasets}

The datasets serve three purposes: supplying training prompts, tracking optimization during
training, and measuring final-checkpoint performance. We describe their task content below;
Appendix~\ref{app:hp} specifies sampling, scoring and harness settings.

\paragraph{Training prompts.}
DAPO-Math-17k supplies competition-style mathematical problems with integer answers
\citep{yu2026dapo}. OpenCodeReasoning provides competitive-programming questions drawn from
multiple programming platforms, together with synthetic reasoning and Python solutions
\citep{ahmad2025opencodereasoning}. Reasoning Gym is a library of procedural problem generators
and task-specific answer verifiers, covering such skills as arithmetic, symbolic manipulation,
logic and games \citep{stojanovski2025reasoninggym}. We use these sources for their
\emph{prompts}: the student generates its own training trajectories, and the routed teacher
provides supervision on those trajectories. Dataset solutions are not imitation targets, and
answer verifiers do not supply a training reward. The pool contains $17{,}856$ prompts,
$5{,}952$ per domain.
Training and evaluation contain no shared problem instances, as checked before training.
The $62$-step training budget uses $1{,}984$ prompt presentations and generates $7{,}936$ responses;
these counts describe the training budget, not the full pool.

\paragraph{Mathematics.}
MATH-500 is a $500$-problem subset of the MATH benchmark
\citep{hendrycks2021math,lightman2023lets}, whose competition problems span algebra, geometry,
number theory and other mathematical topics. Here it serves as an in-training validation set
for training dynamics and the analysis of batching stability (Figures~\ref{fig:curves}a and~\ref{fig:cka}c).
AIME24 and AIME25 contain problems from the 2024 and 2025 American Invitational Mathematics
Examination.
Both examinations use integer-valued answers.\footnote{Official competition descriptions:
\url{https://maa.org/maa-invitational-competitions/}.}
The Minerva suite tests quantitative problem solving \citep{lewkowycz2022minerva}.

\paragraph{Code generation.}
LiveCodeBench collects recent programming-contest problems from LeetCode, AtCoder and
Codeforces and evaluates executable solutions with tests \citep{jain2024livecodebench}.
LCB-e denotes its easy difficulty subset. Appendix~\ref{app:hp} specifies the problem
selection, date window and scoring procedure used in the main tables.
MBPP consists of short, crowdsourced Python programming problems aimed at entry-level
programming skills \citep{austin2021mbpp}. \href{https://github.com/evalplus/evalplus/releases/tag/v0.2.0}{MBPP+} uses EvalPlus's expanded test suites, with dataset
corrections and filtering, to examine correctness under more extensive testing
\citep{liu2023evalplus}.

\begin{figure}[!t]
\centering
\includegraphics[width=\textwidth]{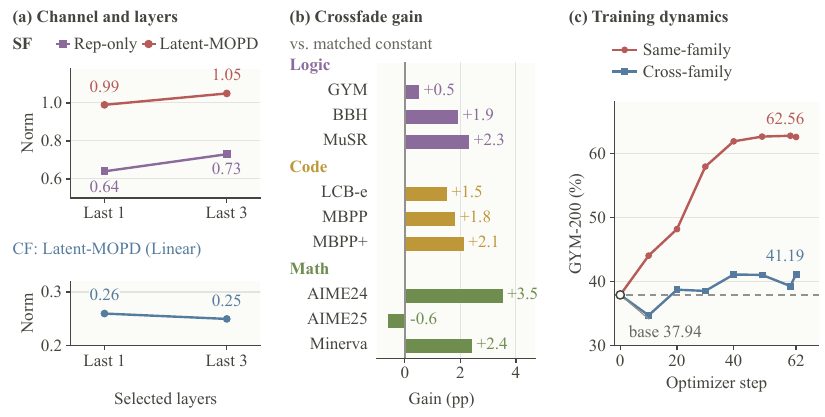}
\caption{\textbf{Layer selection, crossfade and logic-domain training dynamics.}
\textbf{(a)} Last-layer and last-three-layer comparisons: representation-only and
\ours{} in the same-family (SF) setting, and \ours{} with linear projection in the
cross-family (CF) setting. Norm is computed separately within Tables~\ref{tab:s1} and
\ref{tab:s2}; the two regimes have separate axes.
\textbf{(b)} Same-family score gains from crossfade over constant weights, in percentage points,
computed from the reported Table~\ref{tab:s1} scores. Both configurations use last-three-layer
alignment; the comparison changes the coupled token and representation schedule.
\textbf{(c)} \ours{} on GYM-200, with $8$ samples per
problem: same-family last-three-layer and cross-family last-layer linear configurations; the dashed
line marks the base.
The validation set is separate from the $925$-problem GYM evaluation in the main tables.}
\label{fig:additional_analyses}
\end{figure}

\paragraph{Logic and structured reasoning.}
Reasoning Gym evaluates generated answers with each task's own verifier. Our frozen evaluation
set yields $925$ scored problems after the reproduction checks in Appendix~\ref{app:hp}.
We evaluate this set with a $16$k-token generation budget and report its scores as GYM. The batching control (Appendix~\ref{app:detonate}) and Figure~\ref{fig:additional_analyses}c also use GYM-200,
a fixed $200$-problem Reasoning Gym validation set separate from GYM.
BIG-Bench Hard (BBH) collects challenging language and reasoning tasks, including logical
deduction, tracking objects and interpreting structured information \citep{suzgun2022bbh}.
Our harness evaluates $27$ task configurations with $40$ problems each, for $1{,}080$ problems.
MuSR tests multistep reasoning over natural-language narratives, such as drawing conclusions
from the evidence in a murder mystery \citep{sprague2023musr}. It complements procedural
puzzles by requiring the model to connect dispersed statements in a longer narrative.

\paragraph{General ability.}
ARC-Challenge contains challenging multiple-choice grade-school science questions
\citep{clark2018arc}. HellaSwag asks the model to select a plausible continuation of an
everyday situation from competing endings, including adversarially selected distractors
\citep{zellers2019hellaswag}. WinoGrande tests commonsense reference resolution through
sentences with a blank and two candidate completions \citep{sakaguchi2021winogrande}.
None of these three suites supplies prompts to the distillation pool. They probe whether
training on math, code and logic preserves capabilities beyond the trained domains.
We score them by answer-option log likelihood, using length-normalized accuracy for
ARC-Challenge and HellaSwag and accuracy for WinoGrande, not sampled solutions.

\section{Objectives and supervision designs}
\label{app:prelim}
\label{sec:prelim}
\paragraph{On-policy distillation.}
The student $\pi_\theta$ and a frozen teacher $\pi_T$ share one tokenizer with vocabulary $V$.
Given a prompt $x$, the student generates $\hat{y}\sim\pi_\theta(\cdot\mid x)$, and both models then
score every prefix $s_t=(x,\hat{y}_{<t})$ of this response
\citep{agarwal2024onpolicy,gu2024minillm}. Our token channel uses a policy-gradient surrogate
with rewards computed from a no-gradient scoring pass. Let $\bar p_t$ denote the student's
distribution from that pass, $p_{\theta,t}$ its differentiable distribution during the update,
and $p_{T,t}$ the teacher distribution. All three are normalized over the full vocabulary.
We select $V_{k,t}$, the top-$k$ tokens of $\bar p_t$, with $k=16$, and compute, for $v\in V_{k,t}$,
\begin{equation}
\begin{aligned}
w_{t,v} &= \frac{\bar p_t(v)}{\sum_{v'\in V_{k,t}}\bar p_t(v')},\\
r_{t,v} &= -w_{t,v}\bigl[\log\bar p_t(v)-\log p_{T,t}(v)\bigr].
\end{aligned}
\label{eq:token_rewards}
\end{equation}
Thus the candidate weights are normalized within the selected support, while both
log-probabilities in the reward retain their full-vocabulary normalization.

We scale the rewards within each prompt-domain group by their sample standard deviation,
pooling all top-$k$ candidate rewards at valid response positions across that group's
responses in the training batch. With $d$ the prompt's fixed domain, write
$\sigma_d=\max(\operatorname{std}_d(r),10^{-6})$ and $A_{t,v}=r_{t,v}/\sigma_d$.
This scaling does not subtract the mean; groups with fewer than two entries are left unscaled.
Let $M$ index the valid response positions in one
loss calculation, across its participating responses. Writing $\sg$ for stop-gradient,
the token objective is
\begin{equation}
\lopd \;=\; -\frac{1}{|M|}\sum_{t\in M}\sum_{v\in V_{k,t}}
\sg[A_{t,v}]\log p_{\theta,t}(v).
\label{eq:opd}
\end{equation}
The candidate dimension is summed and valid response positions are averaged; gradients flow
only through $p_{\theta,t}$. Appendix~\ref{app:hp} specifies microbatch accumulation and the
schedule weights. Our token-only routed baseline uses this channel without representation
supervision.

\paragraph{Latent supervision.}
We adapt the notation of LastOPD \citep{yang2026lastopd}. At position $t$, let $z^S_t$ and $z^T_t$ be
student and teacher hidden states taken before the LM head, $g_\psi$ a projector from the student
to the teacher width, $\nu(z)=z/\max(\lVert z\rVert_2,\epsilon)$ the normalization, and $\sg$ a
teacher-side stop-gradient:
\begin{equation}
\lrep \;=\; \frac{1}{|M|\,d_T}\sum_{t\in M}
\big\lVert \nu\!\left(g_\psi(z^S_t)\right)-\sg\!\left[\nu(z^T_t)\right]\big\rVert_2^2 .
\label{eq:rep}
\end{equation}
Normalization acts along the hidden-feature dimension. The implementation calls
\texttt{F.normalize} without an explicit epsilon, using its default $\epsilon=10^{-12}$.
This clamps the norm from below; the $10^{-8}$ constant in the masked-mean denominator
(Appendix~\ref{app:hp}) serves a separate purpose.
OPRD \citep{yang2026oprd} matches depth-paired representations directly for compatible models and through frozen
low-rank projector pairs in OPRD-Bridge. LastOPD \citep{yang2026lastopd} applies a normalized hidden-state loss only at the
last-layer state that feeds the head, with a small trainable projector,
and fades out the latent-loss weight over a short crossfade window. Both use one teacher.

\paragraph{Multi-teacher setup.}
We are given a panel of $N$ frozen teachers $\{\pi_{T_i}\}_{i=1}^N$, each an expert of the
student's own capacity or larger, together with a training prompt pool in which every prompt
carries a domain label $d\in\{1,\dots,N\}$ indexing the teacher that owns that domain (math,
code, or logic here). Output-level multi-teacher distillation can form a target by routing
each prompt to $\pi_{T_d}$ (as in MOPD, \citealp{mopd2026}) or by mixing the teachers'
distributions. Our routed token channel uses $p_{T,t}=p_{T_d,t}$ in
Equations~\ref{eq:token_rewards}--\ref{eq:opd}. A probability mixture
$\sum_i\omega_i\,p_{T_i,t}$, with $\omega_i\geq0$ and $\sum_i\omega_i=1$, averages the
probability each teacher assigns to a token. When the
specialist assigns a token more probability than the other teachers do, averaging lowers that
probability. Routing preserves the selected specialist's distribution at each position, and is
the assignment we adopt in both channels. The token objective supervises the student's output distribution; the representation objective also matches selected hidden states before the head.
All teachers contribute across the run through updates to the same student backbone.

\begin{figure}[!t]
\centering
\includegraphics[width=\textwidth]{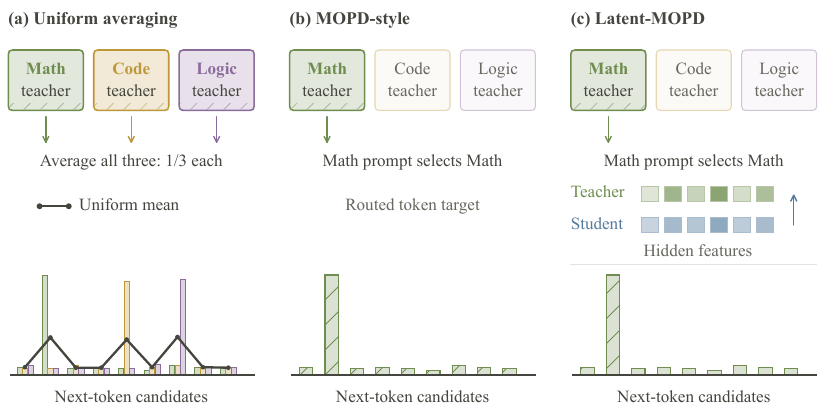}
\caption{\textbf{Three-teacher supervision: an illustrative schematic.} All panels use the same
math prompt. \textbf{(a)} The math, code and logic teachers contribute equally to the averaged
next-token distribution, with weight $1/3$ each. \textbf{(b)} MOPD-style selects the math teacher's
distribution; the code and logic teachers remain inactive. \textbf{(c)} \ours{} retains this token
target and adds a hidden-state target from the same math teacher. Hidden-feature tiles are
separate from the token distributions; their positions do not denote token or layer
correspondence. Values and features are schematic.}
\label{fig:toy}
\end{figure}

Figure~\ref{fig:toy} illustrates the three supervision designs on the same math prompt.

\paragraph{Motivation for a transient latent term.}
LastOPD \citep{yang2026lastopd} studies sustained latent supervision in single-teacher
distillation and motivates supervision at the final prediction interface during a short
crossfade; on its same-lineage pair, which matches our student and same-family math teacher,
keeping the latent term active scores higher on MATH-500 than the crossfade. We nevertheless
start multi-teacher training from the crossfade, giving each specialist its own update clock;
same-family, it outperforms constant weights (Norm $1.05$ vs.\ $0.94$; Table~\ref{tab:s1}). The evidence for our setting comes from the three
multi-teacher settings in Section~\ref{sec:exp} and the controls in Section~\ref{sec:ablation}.
The default recipes use the last three layers same-family and the last layer cross-family.
Tables~\ref{tab:s1}--\ref{tab:s2} compare one versus three layers within each combined crossfade
recipe. Table~\ref{tab:s1} also compares all-layer, last-three-layer and last-layer supervision
under the constant representation-only objective, and includes \ours{} (OPRD-style), an
all-layer variant with constant token and representation weights.

\FloatBarrier

\section{Additional results and channel comparisons}
\label{app:obs}

\paragraph{Same-family: a broad gain that exceeds individual teachers.}
Default \ours{} improves over MOPD-style and all three representation-only baselines on all nine benchmarks,
and exceeds the best teacher on BBH, MuSR, MBPP, MBPP+ and Minerva. Norm reaches $1.05$,
compared with $0.90$ for MOPD-style, $0.64$ for both OPRD-style (all layers) and Rep-only (last 1),
and $0.73$ for Rep-only (last 3).
Under constant all-layer representation supervision, adding the token channel improves
eight of nine benchmarks, raising Norm from $0.64$ to $0.94$ in \ours{} (OPRD-style).
Both configurations use $\lambda=2000$ and the last $2{,}000$ positions.
The default selective crossfade recipe then improves on this joint variant in eight of nine
benchmarks, reaching Norm $1.05$. This comparison evaluates two complete joint configurations,
whose differences are listed in Appendix~\ref{app:hp}.
A separate last-three-layer control keeps other settings fixed and sets both channel weights to one throughout. Crossfade improves eight of nine benchmarks over this
constant-weight control (Figure~\ref{fig:additional_analyses}b), raising Norm from $0.94$ to $1.05$; the control scores higher on
AIME25 ($37.5$ vs.\ $36.9$). This comparison tests the coupled schedule: both the token-weight ramp and representation-weight decay.

Within the same-family crossfade recipe, last-three-layer supervision improves on last-layer supervision in eight of nine benchmarks
(Norm $1.05$ vs.\ $0.99$). Last-layer supervision scores $51.0$ on AIME24 versus $50.8$ for
the default. Both use all valid response positions, $\lambda=1$, the same crossfade schedule
and a $24{,}576$-token packing budget. The representation-only layer controls provide a
second comparison: last-three-layer supervision exceeds both the last-layer and all-layer
alternatives on seven of nine benchmarks each. These runs share the same constant coefficient,
selected positions and packing budget (Appendix~\ref{app:hp}). Thus, the same-family preference
for three late layers appears under both representation-only and joint supervision
(Figure~\ref{fig:additional_analyses}a).

\paragraph{Cross-family: the gain survives a wider teacher panel.}
The default last-layer \ours{} with a linear map improves over MOPD-style on all six reported benchmarks: $+2.9$ and $+1.5$ on AIME24/25,
$+2.5$ and $+1.7$ on MBPP(+), and $+4.9$ and $+3.7$ on GYM and BBH. Logic has the largest
margins, while math and code also improve. On BBH, this variant exceeds representation-only
training by $14.2$ points and token-only MOPD-style by $3.7$ points. The two standalone-channel
comparisons support jointly using token and representation supervision under the tested recipes.
The last-three-layer variant also exceeds both single-channel baselines on all six benchmarks,
with Norm $0.25$ versus $0.26$ for the default last-layer variant. It scores higher on GYM and
MBPP(+), while the default leads on BBH and AIME24/25. Increasing the layer count thus gives
no aggregate improvement in this comparison. Both use the same configured crossfade window
and are evaluated at step $62$.

\paragraph{Parameter-merge initialization: gains after distillation.}
Budget-matched distillation for $62$ steps improves the parameter merge on five of six reported
benchmarks, with gains across math, code and logic. Norm rises from $0.87$ to $1.03$.
The student exceeds the best individual teacher on
BBH, MBPP and MBPP+, three of the six benchmarks. This result shows that
\ours{} can improve a student initialized by parameter merging as well as one initialized from
the base model. All comparisons use the final checkpoints and the evaluation protocol in Appendix~\ref{app:hp}.

\paragraph{Output mixing and routed supervision.}
\label{app:analysis}\label{sec:analysis}
\ours{} improves over uniform averaging on every benchmark reported in Table~\ref{tab:s1},
including GYM (from $40.7$ to $52.4$),
BBH (from $60.0$ to $66.3$), MBPP (from $61.1$ to $65.7$), and MBPP+ (from $50.7$ to $55.7$).
MOPD-style records higher logic scores than uniform averaging, reaching $51.8$ on GYM and $65.4$
on BBH. Both baselines use the same training and evaluation protocol
(Appendix~\ref{app:hp}). In Table~\ref{tab:s1}, the default crossfade recipe also
improves every math, code and logic score over routed token-only MOPD-style.

\paragraph{Logic-domain training dynamics.}
Figure~\ref{fig:additional_analyses}c extends the MATH-500 analysis in Figure~\ref{fig:curves}a
to GYM-200, the fixed $200$-problem validation set. Each checkpoint is evaluated with $8$ samples
per problem, temperature $0.7$, top-$p$ $0.95$ and a $16{,}384$-token generation cap.
From the initial student's $37.94$, the same-family configuration reaches $62.56$ at step
$62$; the cross-family configuration reaches $41.19$. Both stay above the base from step
$20$ onward. These trajectories complement the final-checkpoint results in Tables~\ref{tab:s1}--\ref{tab:s2} by showing how logic performance develops.

\section{Layer-wise representation similarity}
\label{app:cka}

\suppressfloats[t]
\begin{figure}[t]
\centering
\includegraphics[width=0.76\textwidth]{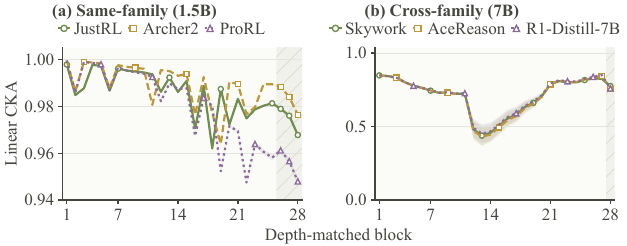}
\caption{\textbf{Depth-matched CKA after masking high-activation dimensions.}
\textbf{(a)} Same-family, with a magnified vertical scale; \textbf{(b)} cross-family, on the
full $0$--$1$ scale. Each teacher is compared with the base
student on the same $192$-prompt diagnostic pool as Figure~\ref{fig:cka}a,b. Before centering,
the top $1\%$ of dimensions by mean absolute activation are zeroed separately in each model
and block. Curves show linear CKA at matched decoder-block outputs, before final normalization;
shading is one standard deviation over $200$ prompt-bootstrap replicates. The masked
cross-family curves retain a middle-layer trough, while the same-family minima occur before
the final block. Hatching marks the selected training depths: the final three blocks in (a)
and the final block in (b).}
\label{fig:cka_masked}
\end{figure}

\paragraph{Models and paired inputs.}
We compare the base DeepSeek-R1-Distill-Qwen-$1.5$B student, before distillation, with all six
teachers (Appendix~\ref{app:hp}) on the same fixed base-student responses. The diagnostic
pool contains $192$ prompts, $64$ per domain. We sample up to $128$ response-token positions
uniformly per response, retaining all positions in shorter responses. Inputs are capped at
$2{,}560$ tokens and positions beyond that cap are excluded, yielding $24{,}568$ paired
positions: $8{,}192$ math, $8{,}189$ code and $8{,}187$ logic. Per-domain comparisons use the
corresponding subsets.

\paragraph{Extraction and CKA.}
We collect decoder-block outputs, numbered $1$--$28$ in every model. The final block is measured
\emph{before} final normalization, whereas distillation targets the state after that normalization
(Section~\ref{sec:method}). Forward passes use bfloat16; extracted features and CKA use float32.
We compare full native feature widths without a learned projector or dimensionality reduction.

For paired activation matrices $X_\ell\in\mathbb{R}^{n\times d_S}$ and
$Y_\ell\in\mathbb{R}^{n\times d_T}$ at matched block $\ell$, let
$\widetilde X_\ell$ and $\widetilde Y_\ell$ denote their column-centered versions.
We compute linear CKA \citep{kornblith2019similarity} as
\begin{equation}
\operatorname{CKA}(X_\ell,Y_\ell)=
\frac{\lVert\widetilde X_\ell^\top\widetilde Y_\ell\rVert_F^2}
{\lVert\widetilde X_\ell^\top\widetilde X_\ell\rVert_F
 \lVert\widetilde Y_\ell^\top\widetilde Y_\ell\rVert_F}.
\end{equation}
Each point pools the selected token rows across domains. Shading gives one empirical standard
deviation over $200$ bootstrap replicates: prompts are sampled with replacement, retaining their
selected token rows. Model checkpoints are fixed; the variation is over diagnostic prompts.

\paragraph{Native profiles.}
Same-family, mean CKA over blocks $1$--$25$ ranges from $0.987$ to $0.992$ across the three
teachers, compared with $0.945$--$0.962$ over the final three blocks. The largest decrease
is at the final block, whose CKA ranges from $0.886$ to $0.930$.
Cross-family, all three native curves reach their minimum at block $13$ ($0.116$--$0.124$),
return to high similarity by block $21$ and remain high through block $27$, and fall again at the final block ($0.536$--$0.635$).
These native profiles motivate the layer choices: the final three layers within the shared lineage,
and the final prediction interface across scales and training lineages, where middle layers differ
substantially. Downstream comparisons evaluate these choices; CKA alone does not determine
which layers are sufficient or optimal for transfer.

\paragraph{Sensitivity to high-activation dimensions.}
We repeat the comparison (Figure~\ref{fig:cka_masked}) after zeroing each model's top $1\%$ of feature dimensions by
mean absolute activation, separately for every block and before centering. This removes
$15$ dimensions in the $1.5$B models and $36$ in the $7$B models. The mask is determined on the
full diagnostic pool and held fixed for domain subsets and bootstrap replicates.
The cross-family trough remains at block $13$, with minima of $0.430$--$0.451$, while the
same-family final-block CKA rises to $0.948$--$0.976$. The masked same-family minima occur at
blocks $18$, $19$ and $22$ for JustRL, Archer2 and ProRL, respectively.
Thus, masking attenuates the native differences and changes where the same-family minima lie;
the cross-family middle-layer trough remains visible under both measurements.

\paragraph{Connection to layer selection.}
The geometry suggests candidate targets; Tables~\ref{tab:s1}--\ref{tab:s2} test their value for
distillation. Same-family, the last three layers outperform the last layer on eight of nine
benchmarks under crossfade and seven of nine under representation-only supervision.
Cross-family, one and three layers split the benchmark wins, with similar aggregate scores
(Norm $0.26$ and $0.25$). The default therefore uses the smaller last-layer target.
Together, the diagnostics and downstream controls support panel-specific layer budgets,
rather than layers chosen solely by CKA minima.

\section{Batch organization and training stability}
\label{app:controls}
\label{app:detonate}
Uniform averaging and interleaved routed targets are distinct designs: the former averages
teacher distributions at each token, while the latter preserves one teacher per prompt and mixes
domains within an optimizer step. In the same-family setting we compared two runs (in-loop validation on MATH-500
and GYM-200, $8$ samples per problem; base student $84.1/37.9$):
the domain-pure reference uses the all-layer representation-only configuration in
Table~\ref{tab:s1}, with $\lambda=2000$, the final $2{,}000$ valid response positions and the
identity map. Both runs use representation-only supervision with \emph{identical} teachers, loss,
coefficient and budget, differing \emph{only} in the batch organization (Figure~\ref{fig:cka}c). With domain-pure blocks of
$32$, training remains stable for $62$ steps ($91.7/56.7$). With the same
$1{:}1{:}1$ rows interleaved, so that one optimizer step averages updates aimed at three different
teachers, both suites fall below the untrained student by step $10$ ($79.8/32.6$) and
performance collapses by step $20$ ($22.8/4.1$), reaching $9.8/0.9$ at step $30$.
At step $62$, performance remains well below the base on both suites ($43.2/12.8$),
showing a lasting deficit within the training budget. Figure~\ref{fig:cka}c shows the
interleaved trajectory over these $62$ steps.
An earlier interleaved run showed the same early collapse ($21.4/4.3$ at step $20$).
In that run, the representation loss \emph{rises} from $7.9\times10^{-5}$ to a peak of
$2.8\times10^{-4}$, without NaNs or gradient overflow in the retained log.
Two additional interleaved runs
reproduce the step-$10$ signature ($82.5/35.8$; $80.7/34.1$), the second also the
step-$20$ collapse ($24.3/4.3$).

Per-prompt routing specifies the
teacher for a prompt and allows either domain-pure or interleaved batches. Accordingly, MOPD's teacher
replacement experiment \citep{mopd2026} and Open-MOPD's analysis of capability imbalance
\citep{gao2026openmopd} address different comparisons from our batching control. Our result
shows that, under the tested protocol, domain-pure updates keep multi-teacher training stable
and reach higher performance within the matched budget.

\begin{figure}[!t]
\centering
\includegraphics[width=\textwidth]{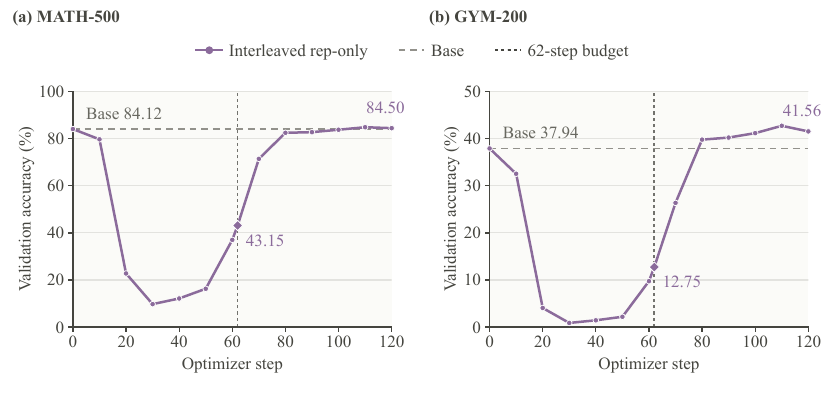}
\caption{\textbf{Interleaved representation-only training under an extended budget.}
The same-family all-layer control in Figure~\ref{fig:cka}c, continued to $120$ steps.
\textbf{(a)} MATH-500; \textbf{(b)} GYM-200. Scores average $8$ samples per problem.
Solid lines connect all measured checkpoints. Horizontal dashed lines mark the base;
vertical dotted lines mark the original $62$-step budget, and diamonds the step-$62$ values. Vertical scales differ.}
\label{fig:mixed_extended}
\end{figure}

\paragraph{The cost of extending interleaved training.}
Figure~\ref{fig:mixed_extended} extends the same interleaved trajectory to $120$ updates.
Even with $1.94\times$ the original budget, MATH-500 ends at $84.50$ (peak $84.90$ at step $110$), close to the
base score of $84.12$, while GYM-200 ends at $41.56$ (peak $42.75$), a modest gain over $37.94$.
Both remain below the domain-pure reference already attained at step $62$ ($91.7/56.7$).
The curves largely level off over the final measured updates: from step $100$ to $120$,
their endpoint gains are only $0.67$ and $0.37$ percentage points, respectively.
Most extra updates offset the collapse, leaving limited gains over the base and a substantial
gap to the shorter domain-pure reference.

\FloatBarrier

\section{Teacher-output alignment after distillation}
\label{app:output_alignment}

We detail Figure~\ref{fig:output_alignment} and extend the comparison across readout depth.
On identical diagnostic prefixes, we compare the final same-family MOPD-style and default
last-three-layer \ours{} checkpoints from Table~\ref{tab:s1} with all three domain teachers.

\paragraph{Inputs and measurement.}
The diagnostic contains $335$ positions: $95$ arithmetic boundaries from $45$ generated chains,
plus three prefix lengths ($50\%$, $70\%$, $90\%$) for each of $40$ code and $40$ logic prompts
from the three-domain pool. Arithmetic prefixes provide preceding correct values and withhold
the current value. All models receive identical prefixes without chat templates.
We compute student--teacher cosine similarity between temperature-one final-head probability
vectors over the shared vocabulary at the last prefix position.
In each domain, we also select positions whose mean pairwise teacher--teacher output cosine is at
or below that domain's median.
This teacher-only mask retains $48$, $60$ and $60$ math, code and logic positions, respectively,
and is fixed across students.

\paragraph{Matching-teacher margin.}
For student $a$, domain $d$ and teacher $s$, let $A_{a,d,s}$ be the mean output-distribution
cosine over the chosen positions in domain $d$. We measure the corresponding teacher's
advantage over the closest alternative teacher as
\[
M_{a,d}=A_{a,d,d}-\max_{s\ne d}A_{a,d,s},
\qquad
\Delta M_d=M_{\text{\ours{}},d}-M_{\text{MOPD-style},d}.
\]
We bootstrap whole chains or prompts within each domain for $3{,}000$ replicates, pairing
all their positions and both students. Each replicate recomputes the margin from
position-weighted means, taking the maximum after averaging. The selection mask stays fixed;
intervals describe variation across diagnostic instances.

\paragraph{Domain-conditioned alignment.}
Both students match their corresponding code and logic teachers most closely
(Figure~\ref{fig:output_alignment}a,b). The math row's maximum shifts from the code teacher
under MOPD-style to the math teacher under \ours{}, raising its margin from $-0.0282$ to $0.0366$.
The paired increase is $0.0648$ ($95\%$ interval $[0.0184,0.1324]$); over all positions it is
$0.0327$ ($[0.0077,0.0659]$). Code and logic increase by $0.0094$ and $0.0013$ on selected
positions, with intervals spanning zero. These diagnostics complement benchmark results by
measuring each recipe's agreement with the domain specialists.

\subsection{Teacher matching across readout depth}
\label{app:depth_matching}

\begin{figure}[H]
\centering
\includegraphics[width=\textwidth]{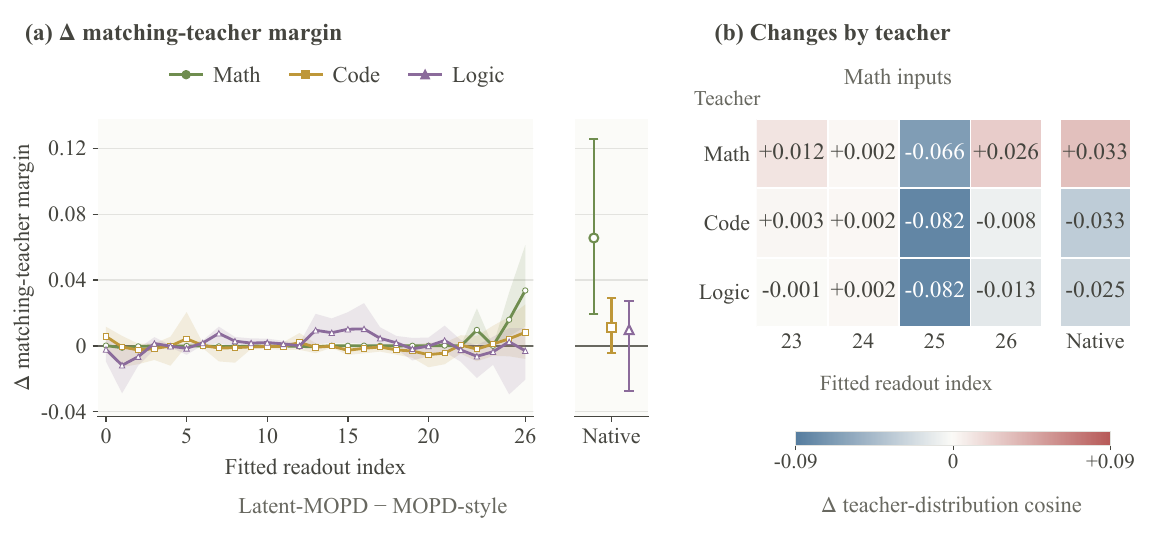}
\caption{\textbf{Teacher matching across readout depth in the same-family setting.}
\textbf{(a)} \ours{} minus MOPD-style matching-teacher margin, by input domain.
Shading and endpoint bars show paired $95\%$ pointwise instance-bootstrap intervals.
\textbf{(b)} Per-teacher cosine changes on selected math positions at the last four fitted
indices and native head; positive values indicate greater similarity under \ours{}.
Both panels use the same teacher-disagreement selection rule. Native-head measurements
are separate from the fitted readouts.}
\label{fig:depth_matching}
\end{figure}

Each student's fitted Jacobian lens \citep{gurnee2026workspace} maps intermediate residuals
to its final residual basis before final normalization and vocabulary readout. On identical
prefixes, we compare temperature-one readouts with all three teachers' final distributions.
Figure~\ref{fig:depth_matching} shows $27$ fitted readout indices and a separately measured native head.

Each student's lens is calibrated independently on the same $150$ prompts spanning math, code
and logic. We retain fitted readout indices $0$--$26$ and apply each readout at the last
position of the diagnostic prefix, evaluating the native head separately.

For this depth diagnostic, we recompute the teacher-only selection from this diagnostic's own teacher
forward passes and hold that mask fixed across students and readout depths. The selected math, code and logic positions cover $35$, $35$ and
$34$ instances, respectively. The margin is computed as above; $3{,}000$ paired instance
resamples are shared across depths and students. At readout index $26$, the math margin
increases by $0.0337$ ($95\%$ interval $[0.0117,0.0612]$); at the native head, it increases by
$0.0655$ ($[0.0196,0.1257]$). Code and logic have smaller native-head changes, with intervals
spanning zero.

Panel~(b) identifies the teacher-specific changes behind the math margin. At index $26$,
similarity to the math teacher increases by $0.026$, while similarity to the code and logic
teachers decreases by $0.008$ and $0.013$. The native head shows the same pattern. The margin
is a relative measure: at index $25$, similarity to all three teachers decreases, but the
decrease is smaller for the math teacher. Thus, the late readouts help locate how the two
training recipes differ in their domain-conditioned agreement with the specialists. The
measurement concerns the distributions exposed by the fitted readouts; it does not establish
that earlier representations lack useful information.

\FloatBarrier

\section{Case studies of generated solutions}
\label{app:cases}

Routed tokens teach \emph{what} to predict; latent alignment additionally supervises the hidden
states used to compute predictions (Figure~\ref{fig:contrast}). We compare MOPD-style and \ours{} on
three examples, using the routed specialists' solutions as references.
Excerpts are verbatim with emphasis added; surrounding text is omitted.

\begingroup
\definecolor{caseframe}{RGB}{218,218,212}
\definecolor{casemopd}{RGB}{240,245,249}
\definecolor{caseours}{RGB}{251,241,235}
\definecolor{caseaccent}{RGB}{164,75,67}
\setlength{\fboxsep}{6pt}
\setlength{\fboxrule}{0.4pt}
\newcommand{\casepair}[3]{%
  \noindent
  \begin{minipage}[t]{0.487\linewidth}
    \fcolorbox{caseframe}{casemopd}{%
      \begin{minipage}[t][#1][t]{\dimexpr\linewidth-2\fboxsep-2\fboxrule\relax}
        \tablefont\raggedright
        \textbf{MOPD-style: token only}\par\smallskip
        #2
      \end{minipage}}
  \end{minipage}\hfill
  \begin{minipage}[t]{0.487\linewidth}
    \fcolorbox{caseframe}{caseours}{%
      \begin{minipage}[t][#1][t]{\dimexpr\linewidth-2\fboxsep-2\fboxrule\relax}
        \tablefont\raggedright
        \textbf{\ours{}: token + latent}\par\smallskip
        #3
      \end{minipage}}
  \end{minipage}\par
}

\noindent\fcolorbox{caseframe}{caseours}{%
  \begin{minipage}{\dimexpr\linewidth-2\fboxsep-2\fboxrule\relax}
    \tablefont
    \textbf{Frozen specialist targets on the student's prefix}\hfill
    \textit{What + how}\par\smallskip
    \begin{minipage}[t]{0.47\linewidth}
      \textbf{Routed token distribution}\par
      $p_{T_d,t}\;\longrightarrow\;p_{\theta,t}$
    \end{minipage}\hfill
    \begin{minipage}[t]{0.51\linewidth}
      \textcolor{caseaccent}{\textbf{Added: teacher hidden-state targets}}\par
      $z^{T_d}_t\;\longrightarrow\;g_\psi(z^S_t)$
    \end{minipage}\par\smallskip
    \ours{} combines both targets through crossfade (Equation~\ref{eq:cross}).
  \end{minipage}}

\paragraph{(a) Code: initializing the computation correctly.}
\textbf{Same-family, MBPP.} Compute the Eulerian number $a(n,m)$; the supplied test is
\texttt{eulerian\_num(3, 1) == 4}. Inner-loop excerpts:

\smallskip
\casepair{48pt}{%
  \ttfamily
  for j in range(i):\\
  \hspace*{1em}if j == 0 or j == i:\\
  \hspace*{2em}euler[i][j] = 0
}{%
  \ttfamily
  for j in range(i):\\
  \hspace*{1em}if j == 0:\\
  \hspace*{2em}\textcolor{caseaccent}{dp[i][j] = 1}
}
\smallskip
\textbf{Computational structure: the recurrence boundary.}
\ours{} sets $a(i,0)=1$ and gets $4$, matching Archer2's closed-form solution.
MOPD-style's zero boundary propagates through the recurrence and returns $0$.

\paragraph{(b) Logic: preserving a state invariant.}
\textbf{Same-family, GYM.} Rewrite until no rule applies:
\[
\mbox{\texttt{A\# \#A}}\to\varnothing,\quad
\mbox{\texttt{B\# \#B}}\to\varnothing,\quad
\mbox{\texttt{A\# \#B}}\to\mbox{\texttt{\#B A\#}},\quad
\mbox{\texttt{B\# \#A}}\to\mbox{\texttt{\#A B\#}}.
\]
The initial state is \texttt{\#B A\# A\# \#A \#B A\# B\# \#A \#A A\#}. Extracted final answers:

\smallskip
\casepair{25pt}{%
  \texttt{\#B\textbackslash{} \#B\textbackslash{} A\#}
}{%
  \texttt{\#B \#B \textcolor{caseaccent}{B\#} A\#}
}
\smallskip
\textbf{Computational structure: valid state transitions.}
The input has three B-tokens, and the rules delete B-tokens only in pairs, so every reachable
state keeps an odd number of them. ProRL and \ours{} match the unique terminal
state; MOPD-style's two-B answer is unreachable even without the backslashes.

\par\noindent\begin{minipage}{\linewidth}
\paragraph{(c) Math: forming the right counting problem.}
\textbf{Cross-family, AIME24.} Count length-$16$ paths across an $8\times8$ grid from lower
left to upper right with four direction changes. Final-solution excerpts:

\smallskip
\casepair{61pt}{%
  ``requires 7 right (R) moves and 7 up (U) moves, totaling 14 moves.''\par\smallskip
  Final answer: $\boxed{180}$.
}{%
  ``Each path consists of \textcolor{caseaccent}{\textbf{8}} right moves (R) and \textcolor{caseaccent}{\textbf{8}} up moves (U), totaling 16 moves.''\par\smallskip
  Final answer: $\boxed{294}$.
}
\par
\end{minipage}\par
\smallskip
\textbf{Computational structure: constraints before counting.}
Skywork and \ours{} use $8$ right and $8$ up moves, giving $2\binom{7}{2}\binom{7}{1}=294$.
MOPD-style's $7+7$ formulation gives $180$ for a length-$14$ path.

\smallskip
With token and representation supervision, \ours{} gets both the answer and the task's key
constraints right in these examples. MOPD-style errs in the recurrence boundary, rewrite invariant
or counting setup. The advantage is visible in the structure of the solutions, illustrating
the motivation for adding latent alignment.
\endgroup

\section{Extended related work}
\label{app:related}

Table~\ref{tab:paradigm} compares representative methods by whether they use multiple teachers
(Multi-t.), hidden-state supervision (Pre-head) and student-generated training samples (On-policy).

\begin{table}[!ht]
\centering
\caption{\textbf{Where multi-teacher capability integration happens.} Among the methods
compared here, \ours{} combines multiple teachers, hidden-state supervision and
student-generated training samples. A dash denotes not applicable.}
\label{tab:paradigm}
\tablefont\setlength{\tabcolsep}{3pt}
\begin{tabular}{lccc}
\toprule
 & Multi-t. & Pre-head & On-policy \\
\midrule
Parameter merging \citep{wortsman2022modelsoups} & \cmark & --- & \xmark \\
Multi-teacher feature KD, off-policy \citep{you2017multiteacher} & \cmark & \cmark & \xmark \\
Uniform averaging & \cmark & \xmark & \cmark \\
MOPD \citep{mopd2026} & \cmark & \xmark & \cmark \\
LastOPD / OPRD \citep{yang2026lastopd,yang2026oprd} & \xmark & \cmark & \cmark \\
\rowcolor{ourrow} \ours{} (ours) & \cmark & \cmark & \cmark \\
\bottomrule
\end{tabular}
\end{table}

\paragraph{On-policy distillation.}
In on-policy distillation, the teacher scores every token of responses that the student samples
itself, reducing the train--inference mismatch of imitating fixed teacher-written text
\citep{agarwal2024onpolicy,gu2024minillm}. It is also used in language-model post-training
alongside supervised fine-tuning and reinforcement learning \citep{yang2025qwen3}. When
teacher and student use different vocabularies, token-level transfer requires handling that
mismatch explicitly \citep{tokbarrier2026}. Our teachers share the student's tokenizer, allowing
us to study representation supervision under the same token-level interface.
Token-level OPD variants differ in which tokens receive a teacher signal and how that signal
is formed, but their targets all come from the teacher's output distribution; the hidden states
that produce those targets receive no direct supervision. \ours{} adds this second source of supervision and routes it across
several domain specialists.

\paragraph{Representation-level distillation.}
Representation matching dates to FitNets \citep{romero2015fitnets}, whose hint layers guide a
narrower student with a teacher's intermediate outputs; later methods align paired hidden states
or attention statistics \citep{sun2019patient,jiao2020tinybert,wang2020minilm,dasgupta2025hidden}. OPRD
\citep{yang2026oprd} brings hidden-state matching to on-policy distillation: along the student's
own rollouts, it matches layers by relative depth and maps unequal
widths through frozen low-rank projectors. OPRD also evaluates joint token and
representation supervision with one teacher and discusses multi-model consolidation as a
potential application. LastOPD \citep{yang2026lastopd} reports
collapse under sustained latent supervision in cross-size experiments, and restricts the latent
term to the final layer, crossfading to token-only OPD over a short window.
Latent-OPD \citep{shen2026latentopd} applies a related trajectory-level latent
signal. PHF \citep{phf2026} and PR-OPD \citep{li2026propdprivilegedrepresentationonpolicy} align hidden-state \emph{flows} and
per-layer states on-policy, against a privileged copy of the student itself. \emph{All of these use a single teacher (or the student
itself) in their experiments.} \ours{} directly studies capability integration with
\emph{multiple} teachers, selecting routed late-layer targets and scheduling their supervision with per-teacher crossfades.
The routed expert provides both hidden-state and token targets;
the default cross-family configuration uses one shared linear map to match the teacher width.

\paragraph{Multi-teacher distillation.}
Multi-teacher distillation trains one student from several teachers. \emph{Off-policy} methods
combine teachers by averaging or adaptively weighting their predictions
\citep{you2017multiteacher,yuan2021reinforced}, while feature-level methods align the student to
\emph{several} teachers' intermediate features or output embeddings on a fixed corpus
\citep{you2017multiteacher,amtmlkd2020,formont2025taskagnostic}; for language models, FuseLLM
\citep{wan2024fusellm} fuses source models through their output distributions. All of these train
on fixed data rather than student-generated trajectories. Recent multi-teacher OPD studies examine domain
routing, scheduling and general-capability retention \citep{mopd2026,d3mopd2026,ucmopd2026}.
Data-efficient OPD also extends to multiple teachers \citep{rethinkopd2026}.
UI-MOPD combines routed token-level distillation with rule-based RL rewards \citep{uimopd2026},
while CA-OPD uses confidence-aware token targets and teacher-guided rollouts for structured
visual prediction \citep{caopd2026}. A comparison of fusion
paradigms studies parameter merging, mixed-domain RL and MOPD \citep{fusionpara2026}.
MT-SDPO \citep{mtsdpo2026} further studies teacher selection, verifying candidates against the
final answer. These approaches concern which output target the student receives. \ours{} adds
selected hidden states from the same routed teacher, extending the supervision available on
each student-generated response.

For image generators, Poly-OPD \citep{fu2026polyopd} combines heterogeneous flow-model teachers
through a pixel bridge and targets in a frozen DINOv2 feature space, while STEP-OPD
\citep{wei2026stepopd} distills task-specialized diffusion teachers on-policy and aligns the
direction and magnitude of student and teacher representation changes across blocks. Our representation channel
matches routed LLM hidden states on student-generated token prefixes.

MOPD \citep{mopd2026} reports three stages: general SFT, domain-specific
RL, and integration. Its distillation stage freezes the teachers. We retain per-prompt routing and on-policy
scoring, reusing existing specialist checkpoints without additional teacher training for this study.
Our token update uses student top-$k$ candidates and fixed advantages derived from teacher--student
log-probability ratios (Section~\ref{sec:method}); MOPD studies a different top-$k$ construction
alongside its policy-gradient estimator. It emphasizes same-origin
teachers and reports degraded transfer after replacing one with a stronger external model.
Our cross-family experiment integrates separately developed $7$B specialists into a $1.5$B student
using both token and latent supervision. These models differ from the student in scale and post-training while
sharing a tokenizer and Qwen backbone. Open-MOPD \citep{gao2026openmopd} examines capability imbalance under domain
routing. Our batching analysis addresses a separate implementation choice: whether individually
routed prompts from several domains contribute to the same optimizer step. Per-prompt routing
is compatible with either batch organization, so the effect of batch organization must be tested directly.

Weight-space merging provides a complementary route to capability integration. Parameter
averaging \citep{wortsman2022modelsoups} and task arithmetic \citep{ilharco2023taskarithmetic}
combine compatible models without distillation. In Section~\ref{sec:s3}, we initialize the
student from a parameter merge for $62$ steps of \ours{} to test whether distillation
can improve on the merged weights.

\paragraph{Representational similarity and diagnostics.}
Centered kernel alignment (CKA) \citep{kornblith2019similarity} compares representations across
networks with different widths. We compare each specialist with the base student at matched
block depths on shared student-generated responses (Figure~\ref{fig:cka}a,b). The same-family
profiles remain close through most blocks, with a larger native-CKA drop at the final block;
the cross-family profiles have a pronounced middle-layer trough and recover in later blocks.
High-activation dimensions \citep{sun2024massive} can affect representational similarity
\citep{yang2026lastopd}; Appendix~\ref{app:cka} specifies the measurement and
examines sensitivity to masking those dimensions. These diagnostics describe representation
geometry before distillation. CKA alone does not establish which layers suffice for transfer
or identify the computations responsible for downstream gains.

\section{Scope and extensions}
\label{app:limits}

\paragraph{Evaluation setting.}
The experiments compare ways to integrate a fixed set of specialists: output mixing, routed
token supervision, representation-only supervision, our combined objective, and parameter merging.
Appendix~\ref{app:hp} documents the protocol and the implementation of our MOPD-style baseline.
The main tables evaluate the final checkpoints. We use fixed domain labels and set the
representation-loss scale for each setting. All teachers share the student's tokenizer. More diverse teachers, learned
routing and support for different tokenizers would extend the setting studied here.

\paragraph{Channel and design comparisons.}
Tables~\ref{tab:s1}--\ref{tab:s2} test token supervision alone, representation supervision alone,
and their combination in \ours{}. These comparisons establish the advantage of the joint
method over either standalone channel under the reported configurations. MOPD-style uses a constant
token-loss weight, representation-only training retains the latent term throughout, and the
default joint method uses crossfade. The same-family constant-weight last-three-layer joint
control holds the other settings fixed, testing the crossfade schedule as a whole.
Within each regime, the last-layer and last-three-layer
\ours{} variants hold other settings fixed to assess layer selection. The batching control
(Section~\ref{sec:detonate}) holds per-prompt routing fixed and tests how grouping domains
within an optimizer step affects stability.

\paragraph{Representation evidence and extensions.}
The \emph{what/how} framing identifies the two supervision targets: token distributions and
the hidden states used to compute them. The case studies (Appendix~\ref{app:cases}) examine computational structure
in generated outputs, while depth-matched CKA describes how each teacher's representations
resemble the base student's before training (Appendix~\ref{app:cka}). The native and masked
profiles show how these comparisons depend on layer and activation dimensions. Measurements
of learned projectors and update directions at the states used as training targets (after final normalization, unlike the CKA readouts) would
help connect this geometry to the improvements from representation supervision.

\end{document}